\documentclass[10pt]{article} 
\usepackage{amsthm}
\usepackage{amsmath}
\usepackage{amssymb}
\usepackage[preprint]{tmlr}

\usepackage{amsmath,amsfonts,bm}

\def\eqref#1{equation~\ref{#1}}

\def\1{\bm{1}}

\DeclareMathAlphabet{\mathsfit}{\encodingdefault}{\sfdefault}{m}{sl}
\SetMathAlphabet{\mathsfit}{bold}{\encodingdefault}{\sfdefault}{bx}{n}

\usepackage{graphicx}
\usepackage{hyperref}
\usepackage{url}
\usepackage{subcaption}
\usepackage{caption}
\usepackage{booktabs}
\usepackage{multirow}

\title{$\bm{\lambda}$-VAE: Variance Equalization for Posterior Collapse}

\author{\name Girum Demisse \email girumdemisse@microsoft.com \\
      \addr Microsoft African Development Center\\
      Nairobi Kenya
      }

\newtheorem{prop}{Proposition}

\def\month{MM}  
\def\year{YYYY} 
\def\openreview{\url{https://openreview.net/forum?id=XXXX}} 

\begin{document}

\maketitle
 
\begin{abstract}
Variational Autoencoders (VAEs) frequently suffer from posterior collapse, a
failure mode in which the approximate posterior converges to the prior,
rendering the latent code uninformative. Despite extensive research, a unified
account of why collapse occurs has remained an open question. We identify and
formalize two logically independent but coupled causes. \emph{Gradient
imbalance} occurs when the decoder's reconstruction signal vanishes faster than
the $\mathbb{KL}$ regularization pressure as the posterior widens.
\emph{Information gap} occurs when the stochastic sampling step discards a
substantial fraction of the encoder's computed representation, attenuating
decoder sensitivity and making collapse inexpensive. Both causes share the same
collapse trajectory, and we show that the information gap is algebraically
equivalent to mismatch between the aggregate posterior and the prior, unifying
two pathologies. Subsequently, we introduce $\lambda$-VAE, which resolves both
causes through a single modification to the reparameterization step: the
sampling noise is scaled by per-dimension exponent, while the $\mathbb{KL}$
penalty retains the original posterior variance. This asymmetry shifts the
stable training attractor away from the degenerate collapsed state, driving all
latent dimensions toward the same equilibrium -- a mechanism we term
\emph{variance equalization}. A closed-form optimal exponent per dimension
follows from a net information gain objective, with a single hyperparameter
controlling the reconstruction--generation tradeoff. We validate on standard
benchmarks (Binary MNIST, Binary Omniglot, CIFAR-10, CelebA-64), showing
consistent reductions in collapsed dimensions, information capacity gains of up
to $2.8\times$ nats, and reconstruction quality improvements of up to $+0.33$ BPD.
\end{abstract}


 
\section{Introduction}
\label{sec:intro}
 
Variational Autoencoders (VAEs)~\cite{kingma2013auto,rezende2014stochastic} have
become a foundational tool for learning latent representations, with
applications in image generation~\cite{razavi2019generating}, natural language
processing~\cite{bowman2015generating}, multimodal
modelling~\cite{shi2019variational}, and semi-supervised
learning~\cite{kingma2014semi}. VAE is one of many variational inference
frameworks that use approximate posterior for maximizing data
likelihood~\cite{neal1998view,mclachlan2007algorithm,minka2013expectation,
jordan1999introduction}. It is, however, one of the first to use
\emph{amortization} to estimate approximate posterior; an encoder learns to map
each input to a posterior distribution over a latent space, and a decoder
reconstructs the input from a sample. This is a highly efficient and scalable
approach, but it is prone to a well-documented failure mode called
\emph{posterior collapse}~\cite{bowman2015generating, chen2016variational,
lucas2019don}, in which the learned approximate posterior converges to the prior
partially or fully. When collapse occurs, the decoder learns to reconstruct data
from the prior. In its extreme form, the mutual information between data and
latent variables goes to zero ($I(X;Z) \approx 0$), and the latent space loses
all its representational value. This failure is most severe with high-capacity
decoders~\cite{bowman2015generating, razavi2019generating}, such as those with
autoregressive or deep convolutional architectures, but it occurs to varying
degrees in most practical VAE training runs.
 
Extensive research has proposed different solutions for posterior collapse;
re-weighting the Kullback-Leibler ($\mathbb{KL}$) divergence
term~\cite{higgins2016beta,fu2019cyclical}, imposing minimum $\mathbb{KL}$
floors~\cite{chen2016variational,razavi2019preventing}, matching the aggregate
posterior explicitly~\cite{zhao2019infovae,tolstikhin2017wasserstein}, enriching
the posterior family~\cite{rezende2015variational,kingma2016improved}, and
adjusting training dynamics~\cite{he2019lagging}. These methods target different
manifestations of the problem and share a common limitation: they act on the
training objective and propose a global solution, while lacking a unified
account on the causes of collapse.

 
In this paper, we provide a formal account of why posterior collapse occurs and
build a targeted solution on top of it. Our starting observation is simple: in a
standard VAE, the $\mathbb{KL}$ term selectively pushes low-signal dimensions
toward $\sigma_i = 1$ while leaving high-signal dimensions active, producing a
\emph{polarized} distribution of posterior variances. This polarization is both
the signature of emerging collapse and a driver of further collapse. Hence, the
right intervention is not to globally modify the training objective but to
counteract the polarization directly per dimension, which we formalize and
validate in subsequent sections.

\noindent\textbf{Summary of Contributions:}
\begin{enumerate}
    \item 
  We prove that posterior collapse has two logically independent but
  coupled causes: \emph{gradient imbalance}
  (Proposition~\ref{prop:ratio_grad}), in which the reconstruction
  gradient vanishes before the $\mathbb{KL}$ restoring force as the posterior
  widens, and \emph{information gap} (Proposition~\ref{prop:enabling}),
  in which the stochastic bottleneck discards encoder signal, making
  collapse inexpensive. We further show that the information gap is
  algebraically equivalent to marginal mismatch (difference between the aggregate
  posterior and the prior) (Eq.~\ref{eq:duality}).

    \item 
  We propose $\lambda$-VAE, replacing the reparameterization noise
  $\sigma\epsilon$ with $\sigma^\lambda\epsilon$ while retaining the
  $\mathbb{KL}$ penalty on the original $\sigma^2$ resolving both causes
  simultaneously. The asymmetry drives all latent dimensions toward the same
  equilibrium -- \emph{variance equalization} -- without modifying the training
  objective or adding parameters. A closed-form optimal per-dimension exponent
  follows from a net information gain objective
  (Proposition~\ref{prop:optimal_lambda}), with a single hyperparameter $\delta$
  controlling the reconstruction--generation tradeoff.

    \item 
  We show that on binarized benchmarks, $\lambda$-VAE reduces collapsed
  dimensions from 16 to 1 on Binary MNIST and 13 to 0 on Binary Omniglot while
  improving reconstruction quality. On RGB images, information capacity grows by
  up to $2.8\times$ and BPD improves by $+0.33$ on CIFAR-10. A PixelCNN
  experiment reveals a fundamental limitation of BPD as a collapse diagnostic:
  two models with near-identical BPD (3.518 vs.\ 3.494) differ by $6.2\times$ in
  the decoder capacity allocated to the latent code.
\end{enumerate}
 
 
\section{Variational Autoencoders}
\label{sec:background}
 
VAE~\cite{kingma2013auto,rezende2014stochastic} is one
of the earlier approaches to use \emph{amortization} and stochastic gradient
estimation in large-scale datasets via the Evidence Lower Bound (ELBO), written as
\begin{align}
    \mathcal{L}
    = \mathbb{E}_{q_\psi(z|x)}\!\left[\log p_\theta(x|z)\right]
    - \mathbb{KL}\!\left(q_\psi(z|x)\,\|\,p(z)\right),
    \label{eq:elbo}
\end{align}
with $q_\psi(z|x) = \mathcal{N}(\mu(x), \mathrm{diag}(\sigma^2(x)))$, $p(z) =
\mathcal{N}(0, \mathbf{I})$, and sampling via $z = \mu(x) + \sigma(x)\epsilon$, $\epsilon
\sim \mathcal{N}(0,\mathbf{I})$. The Gaussian assumption on the posterior and prior makes
the $\mathbb{KL}$ term computable in a closed form; $\mathbb{KL}_i =
\frac{1}{2}(\mu_i^2 + \sigma_i^2 - \log\sigma_i^2 - 1)$, henceforth we work per
latent dimension $i$ and suppress the subscript where unambiguous.
In~\cite{hoffman2016elbo}, the $\mathbb{KL}$ term in ELBO is decomposed and the 
overall loss is written as
\begin{align}
    \mathcal{L}
    = \mathbb{E}\!\left[\log p_\theta(x|z)\right] - I_q(X;Z)
    - \mathbb{KL}(q_\psi(z)\|p(z)),
    \label{eq:decomp}
\end{align}
where $I_q(X;Z)$ is the mutual information, and $\mathbb{KL}(q_\psi(z)\|p(z))$
is the marginal mismatch. Hence, minimizing the $\mathbb{KL}$ term in
Eq.~\ref{eq:elbo} reduces both $I_q(X;Z)$ and the marginal mismatch; the
optimizer cannot distinguish between them, which is why collapse is an
unintended consequence. That is, at $\mathbb{KL} \approx 0$ in
Eq.~\ref{eq:elbo}, the latent code $z$ becomes statistically independent of the
input $x$, rendering the encoder information useless in the signal
reconstruction.

In general, ELBO creates an information bottleneck that encourages encoding the
least amount of information subject to reconstruction
error~\cite{tishby2000information}. Posterior collapse, however, is an extreme
case in which the model achieves sufficient reconstruction accuracy while
ignoring all input signal, $I_{q}(X;Z) \approx 0$ -- often observed and measured
per latent dimension. In Section~\ref{sec:posterior_collapse}, we will expand on
both data and model specific conditions that enable VAE training to find
solutions that fail to encode details about the input while achieving high
reconstruction accuracy.

 
 
\section{Posterior Collapse: Two Causes}
\label{sec:posterior_collapse}
 
Posterior collapse in trained models manifests as near-zero per-dimension
$\mathbb{KL}$, a low count of active latent dimensions, and $I_q(X;Z) \approx
0$. In the subsequent sections, we present two principal causes for
posterior collapse and show their relationship to $\mathbb{KL}$ and mutual
information.
 
\subsection{Gradient Imbalance}
\label{subsec:grad_imbalance}
 
Differentiating Eq.~\ref{eq:elbo} with respect to $\sigma_i$ via the
reparameterisation $z_i = \mu_i + \sigma_i\epsilon_i$, we have
\begin{align}
    \frac{\partial\mathcal{L}}{\partial\sigma_i}
    =  \mathbb{E}_{p(x),\,\epsilon_i}\!
        \bigl[\nabla_{z_i}\log p_\theta(x|z)\cdot\epsilon_i\bigr]
    + \frac{1}{\sigma_i} - \sigma_i.
    \label{eq:gradient_sigma}
\end{align}
The $\mathbb{KL}$ restoring force (second term) is positive for $\sigma_i < 1$,
zero at $\sigma_i = 1$, and negative for $\sigma_i > 1$. As such, it always
drives $\sigma_i$ toward 1 while looking for a stable solution
in Eq.~\ref{eq:gradient_sigma}. When the first term, the reconstruction gradient
which we denote by $g_i^\sigma$, vanishes then $\sigma_i = 1$ is the sole
equilibrium. Hence, the dynamics between $g_i^\sigma$ and the $\mathbb{KL}$
restoring force determine whether a dimension remains active or collapses during
training.

We define the \emph{gradient ratio} $\rho_i = |g_i^\sigma| / |1/\sigma_i -
\sigma_i|$ as the primary indicator of collapse. That is, when $\rho_i \ll 1$,
the $\mathbb{KL}$ term dominates and the dimension collapses. The following
proposition formalizes this understanding.
 
\begin{figure}[t]
\centering
\begin{subfigure}[t]{0.45\linewidth}
  \centering
  \includegraphics[width=\linewidth]{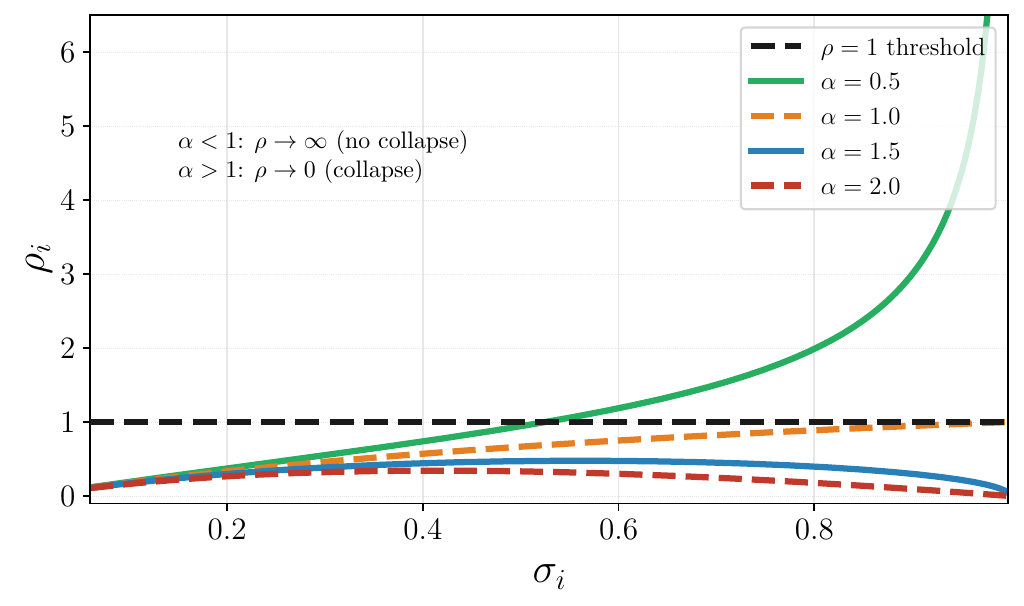}
  \caption{Gradient imbalance ($\rho_i$ vs $\sigma_i$)}
  \label{fig:rho_curve}
\end{subfigure}
\hspace{0.04\linewidth}
\begin{subfigure}[t]{0.42\linewidth}
  \centering
  \includegraphics[width=\linewidth]{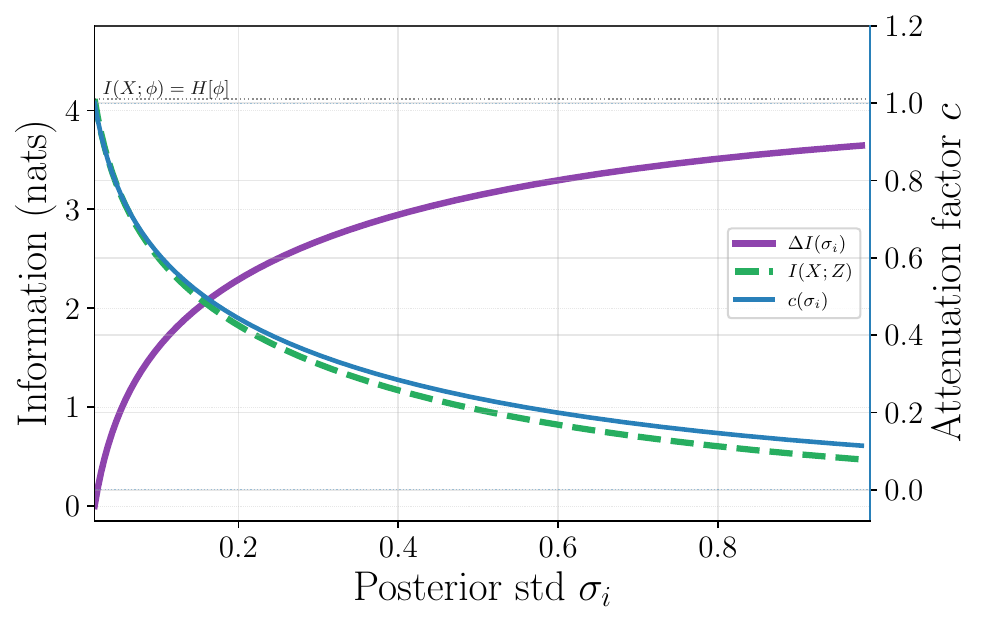}
  \caption{Information gap ($\Delta I$, $I(X;Z)$, $c$ vs $\sigma_i$)}
  \label{fig:gap_curve}
\end{subfigure}
\caption{\textbf{Posterior collapse causes.}
\emph{(a)} Gradient ratio $\rho_i$ under the decay model $g_i^\sigma =
g_0(1-\sigma_i)^\alpha$, $g_0 = 2$. For $\alpha > 1$
(Proposition~\ref{prop:ratio_grad}), $\rho_i \to 0$ as $\sigma_i \to 1$: the
reconstruction signal vanishes faster than the $\mathbb{KL}$ restoring force,
making the collapsed state the only stable fixed point. For $\alpha < 1$,
$\rho_i$ is no longer a reliable indicator: it stays above 1 even when the
dimension has collapsed, because the reconstruction gradient from the decoder
decays much more slowly in this regime. \emph{(b)} Mutual information $I(X;Z)$,
information gap $\Delta I$, and attenuation factor $c = I(X;Z)/I(X;\phi)$ using
the Gaussian channel model with $S_i = 1.5$. As $\sigma_i$ increases towards 1:
$I(X;Z)$ falls, $\Delta I$ grows toward $I(X;\phi)$, and $c \to 0$. Hence, collapse
becomes much cheaper (Proposition~\ref{prop:enabling}).}
\label{fig:pathologies}
\end{figure}

\begin{prop}[Gradient Imbalance]
\label{prop:ratio_grad}
Suppose the asymptotic decay condition for the reconstruction gradient is
$|g_i^\sigma| = O((1-\sigma_i)^\alpha)$ as $\sigma_i \to 1^-$ with $\alpha > 1$.
Then dimension $i$ collapses to the prior if and only if $\liminf_{t \to
\infty}\rho_i(t) < \rho_{\mathrm{crit}}$ for some finite $\rho_{\mathrm{crit}} >
0$.%
\footnote{The idealized gradient-balance value is $\rho_{\mathrm{crit}}=1$ (any
non-collapsed equilibrium of $\partial\mathcal{L}/\partial\sigma_i$ satisfies
$\rho_i \geq 1$). In practice, joint encoder–decoder optimization shifts the
effective threshold; on CIFAR-10, for instance, the empirical
$\rho_{\mathrm{crit}}\approx 0.006$ separates active from collapsed dimensions
with 99.8\% accuracy. The value 1 therefore bounds the theoretical regime; the
operative threshold is model and dataset specific.}
\end{prop}
 
\noindent
Under this condition, $\rho_i \to 0$ as $\sigma_i \to 1$, making $\sigma_i = 1$
the sole stable fixed point. The decay exponent $\alpha$ characterizes the
decoder's sensitivity to the latent code near collapse. The condition $\alpha >
1$ identifies decoders for which $\rho_i$ is a necessary and sufficient collapse
signal.
Autoregressive decoders satisfy this with $\alpha \gg 1$, explaining their
high collapse rate~\cite{bowman2015generating} and empirically confirmed by the
PixelCNN experiment in Section~\ref{sec:exp_generation}.
The collapsed fixed point $\sigma_i^* = 1$ is locally stable with linearization
rate $-2$; by contrast, $\mu_i^* = 0$ has rate $-1$, which is why $\sigma_i \to
1$ is the more reliable leading indicator of collapse. Full proof and details
are in Appendix~\ref{ap:gradient_imbalance}, while illustration of the dynamics
is shown in Figure~\ref{fig:rho_curve}.
 
\subsection{Information Gap}
\label{subsec:info_gap}
 
The encoder in a VAE, which estimates the posterior parameters, is a
deterministic learned map $\phi(x) = (\mu(x), \log\sigma(x))$, while the
latent $Z$ is obtained by stochastic reparameterization. This defines a Markov
chain $X \to \phi(X) \to Z$, for which the data processing
inequality~\cite{cover1999elements} gives:
\begin{align}
    H[X] \;\geq\; I(X;\phi) \;\geq\; I(X;Z) \;\geq\; 0,
    \label{eq:hierarchy}
\end{align}
where $H(\cdot)$ denotes entropy of a random variable. Since $I(X;\phi) =
H[\phi]$ (encoder is deterministic), the \emph{information gap} is the
information capacity that the sampling step loses, captured as
\begin{align}
    \Delta I \;=\; H[\phi] - I(X;Z) \;\geq\; 0.
    \label{eq:gap}
\end{align}
The information gap widens if the dataset has low signal power (low variance in
$\mu_i(X)$), which reduces $I(X;Z)$ as $\sigma \to 1$. More importantly,
a large gap implies an attenuation factor that dampens decoder sensitivity to
encoder change, facilitating collapse. The following proposition formalizes
this.
\begin{prop}[Information Gap]
\label{prop:enabling}
Let $\mathcal{R} = f(I(X;Z))$ be reconstruction quality measure for
monotonically increasing $f$. The sensitivity of reconstruction quality to
encoder capacity is attenuated as
\begin{align}
    \frac{\partial\mathcal{R}}{\partial I(X;\phi)}
    = \frac{\partial\mathcal{R}}{\partial I(X;Z)}
    \cdot \left(1 - \frac{\partial\Delta I}{\partial I(X;\phi)}\right).
    \label{eq:enabling}
\end{align}
\end{prop}
 
\noindent
The attenuation factor $c = 1 - \partial\Delta I/\partial I(X;\phi) \approx
I(X;Z)/I(X;\phi)$ (under the approximation that the two move proportionally
during training) measures how much of the encoder's signal the decoder can
access. 

The Gaussian channel model makes the information gap view concrete by
treating $z_i = \mu_i(X) + \sigma_i\epsilon_i$ as transmitting signal $\mu_i$
through noise $\sigma_i\epsilon_i$, giving
\begin{align}
    I(X;Z_i) \;\approx\; \tfrac{1}{2}\log\!\left(1 + \frac{S_i}{\sigma_i^2}\right),
    \label{eq:gaussian_channel}
\end{align}
where $S_i = \mathrm{Var}[\mu_i(X)]$ is the signal power. High SNR
(signal-to-noise ratio) gives $c \approx 1$, making collapse costly; low SNR
gives $c \to 0$ and collapse exerts zero pressure on the decoder -- it is free
(Figure~\ref{fig:gap_curve}). More importantly, under marginal matching where
$q(z) \approx \mathcal{N}(0, \mathbf{I})$ the signal power $S_i \approx 1 -
\sigma_i^2$, this reduces to $I(X;Z_i) \approx -\log\sigma_i$ as a bound to the
accessible information which goes to zero as $\sigma \to 1$; result is further
discussed in Appendix~\ref{app:information_gap}.

Overall, the information gap decomposes into a reducible and an irreducible
part. The irreducible floor is set by the data's intrinsic dimensionality:
dimensions where the data provides no signal which contribute an unavoidable gap
regardless of encoder quality. For instance, in binarized MNIST and Omniglot,
corner pixels carry no information; when encoded into a high-dimensional latent
space, the corresponding dimensions will naturally collapse. The reducible part
is $\mathbb{KL}$-induced; since $\mathbb{KL}$ forces $S_i \approx 1 -
\sigma_i^2$, any rise in $\sigma_i$ under regularization pressure directly
suppresses $S_i$, creating an information gap a higher-SNR encoder could avoid.
Intuitively, a lower information gap implies a more distributed representation
-- all dimensions carry useful signal, ideally independent rather than redundant --
whereas a sparse representation concentrates signal in few overloaded dimensions
and leaves the rest high-noise, see~\cite{hinton1986learning,bengio2013representation}. 

A direct consequence follows from the ELBO decomposition~\cite{hoffman2016elbo}.
Substituting $I(X;Z) = I(X;\phi) - \Delta I$ into Eq.~\ref{eq:decomp}:
\begin{align}
    \mathbb{KL}(q_\psi(z)\|p(z)) = \mathbb{KL}(q_\psi(z|x)\|p(z)) - I(X;\phi) + \Delta I.
    \label{eq:duality}
\end{align}
This identity holds at every training snapshot. Any increase in $\Delta I$
that is not compensated by a decrease in $\mathbb{KL}(q(z|x)\|p(z)) - I(X;\phi)$
manifests directly as growing marginal mismatch. Encoder attenuation and
aggregate posterior misalignment are therefore algebraically equivalent: two
descriptions of the same quantity, one from the sampling bottleneck and one from
the aggregate distribution. Hence, the information gap serves as a cheap,
per-batch proxy for marginal mismatch.
 
\subsection{Coupling Between the Two Causes}
\label{subsec:coupling}
 
The two causes are logically independent; gradient imbalance depends on decoder
architecture ($\alpha$, and decoder capacity); the information gap depends on
data complexity ($S_i$) and posterior width ($\sigma_i$). Either can occur
without the other. However, both are driven by $\sigma_i \to 1$: gradient
imbalance makes this trajectory dynamically inevitable once $\rho_i <
\rho_{\mathrm{crit}}$ under the asymptotic condition $\alpha > 1$; the
information gap makes reaching $\sigma_i = 1$ less costly before that point.

Once gradient imbalance pushes $\rho_i$  below the critical point, $\sigma_i$
begins to rise. For data with low latent complexity $S_i$, the rising $\sigma_i$
opens a substantial information gap, lowers the attenuation factor $c$, and
weakens the reconstruction gradient $g_i^\sigma$ which reduces $\rho_i$ further
and accelerates the rise in $\sigma_i$. The feedback cycle is closed and
positive; gradient imbalance provides the initial push, and the information gap
amplifies it when the encoder has high capacity. On complex data, the
information gap remains shallow and the attenuation is weak; collapse is then
driven primarily by gradient imbalance alone. In both cases, however, the
polarization of $\sigma$ is the signature of the trajectory toward collapse, and
the key point of intervention. In Section~\ref{sec:lambda_vae}, we present an
approach that actively counteracts this polarization, addressing both causes
simultaneously.
 
 
\section{$\lambda$-VAE: Variance Equalization}
\label{sec:lambda_vae}
 
 
Consider the modification of the latent sampling step in ELBO as
\begin{align}
    z = \mu(x) + \sigma(x)^{\lambda}\cdot\epsilon, \qquad \lambda \geq 1,
    \label{eq:lambda_rep}
\end{align}
while computing the $\mathbb{KL}$ term using the original variance
$\sigma(x)^2$. Subsequently, the objective will be modified as
\begin{align}
    \mathcal{L}_\lambda
    = \mathbb{E}_{q_\lambda(z|x)}\!\left[\log p_\theta(x|z)\right]
    - \mathbb{KL}\!\left(q_\psi(z|x)\,\|\,p(z)\right),
    \label{eq:lambda_elbo}
\end{align}
where $q_\lambda(z|x) = \mathcal{N}(\mu(x),
\mathrm{diag}(\sigma(x)^{2\lambda}))$ is the sampling distribution but the
$\mathbb{KL}$ penalty uses $\sigma(x)^2$ and not $\sigma(x)^{2\lambda}$; this is
$\lambda$-VAE.

$\lambda$-VAE introduces an asymmetry between the sampling and penalty terms.
The decoder receives noise $\sigma_i^{\lambda}\epsilon$ (reduced relative to
$\sigma_i\epsilon$ for $\sigma_i < 1$, $\lambda > 1$) while the $\mathbb{KL}$
penalty evaluates the original $\sigma_i^2$. This decoupling allows more
information to flow through the bottleneck without inflating the $\mathbb{KL}$
penalty. 

 
\subsection{Gradient Rebalancing}
\label{subsec:grad_rebalancing}
 
The ELBO gradient with respect to $\sigma_i$ under $\lambda$-scaling is
\begin{align}
    \frac{\partial\mathcal{L}_\lambda}{\partial\sigma_i}
    = \lambda\,\sigma_i^{\lambda-1}\cdot g_i^\sigma
    + \frac{1}{\sigma_i} - \sigma_i.
    \label{eq:lambda_gradient}
\end{align}
Hence, the reconstruction term is amplified by $\lambda\sigma_i^{\lambda-1}$. As
a result, the
effective gradient ratio becomes $\rho_i^\lambda = \lambda\sigma_i^{\lambda-1}\rho_i$.
As $\sigma_i \to 1$, $\rho_i^\lambda \to \lambda\rho_i$. A dimension with
$\rho_i < \rho_{\mathrm{crit}}$ (which would collapse under $\lambda = 1$) is protected once
$\lambda > \rho_{\mathrm{crit}}/\rho_i$.

Furthermore, a given $\lambda_i$ defines an attractor $\sigma_i$ at which point
the amplification vanishes. The mechanism drives $\sigma$ above or below the
stable point towards it; see Figure~\ref{fig:equalization_convergence} and
Appendix~\ref{app:sigma_star}. This behaviour leads to variance equalization
across dimensions, which is the key mechanism of $\lambda$-VAE to counteract
\emph{polarization} and hence posterior collapse.
 
\subsection{Information Gain}
\label{subsec:info_gain}
 
Under $\lambda$-scaling, using $H[Z] \approx H[Z_\lambda]$ (valid for moderate
$\lambda$; Appendix~\ref{app:information_gap}), we have the following
information gain
\begin{align}
    I(X; Z_\lambda) - I(X; Z_1)
    = \frac{\lambda-1}{2}\sum_{i=1}^d \mathbb{E}\!\left[-\log\sigma_i^2\right]
    \geq 0.
    \label{eq:info_gain}
\end{align}
All terms are positive for $\sigma_i < 1$, giving a residual information gap
$\Delta I_\lambda = \Delta I_1 \cdot (2-\lambda)$. As such, for a given training
snapshot, at $\lambda = 2$ the gap is eliminated, and for $\lambda \in (1,2)$ it
is partially closed. The information gain per dimension is proportional to
$\mathbb{E}[-\log\sigma_i^2]$, and thus confident dimensions (small $\sigma_i$)
yield larger information gains than dimensions with wide posterior.
 
\subsection{Variance Equalization and the Equilibrium Target}
\label{subsec:equalization}
 
\begin{figure}[t]
\begin{subfigure}[t]{0.45\linewidth}
  \centering
  \includegraphics[width=\linewidth]{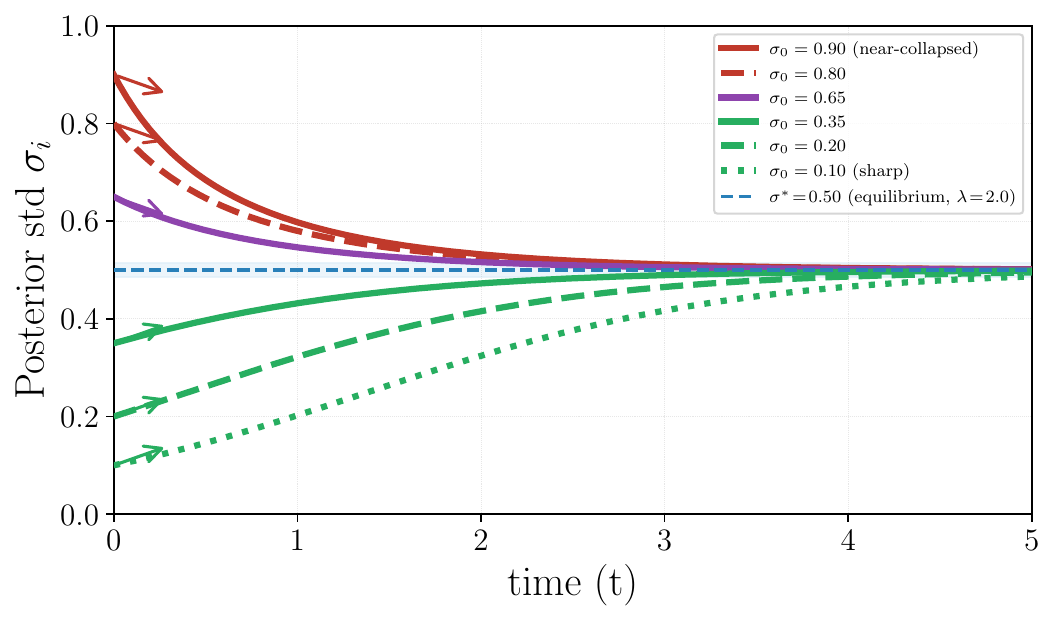}
  \caption{Convergence to $\sigma^*$}
  \label{fig:equalization_convergence}
\end{subfigure}
\hfill
\begin{subfigure}[t]{0.45\linewidth}
  \centering
  \includegraphics[width=\linewidth]{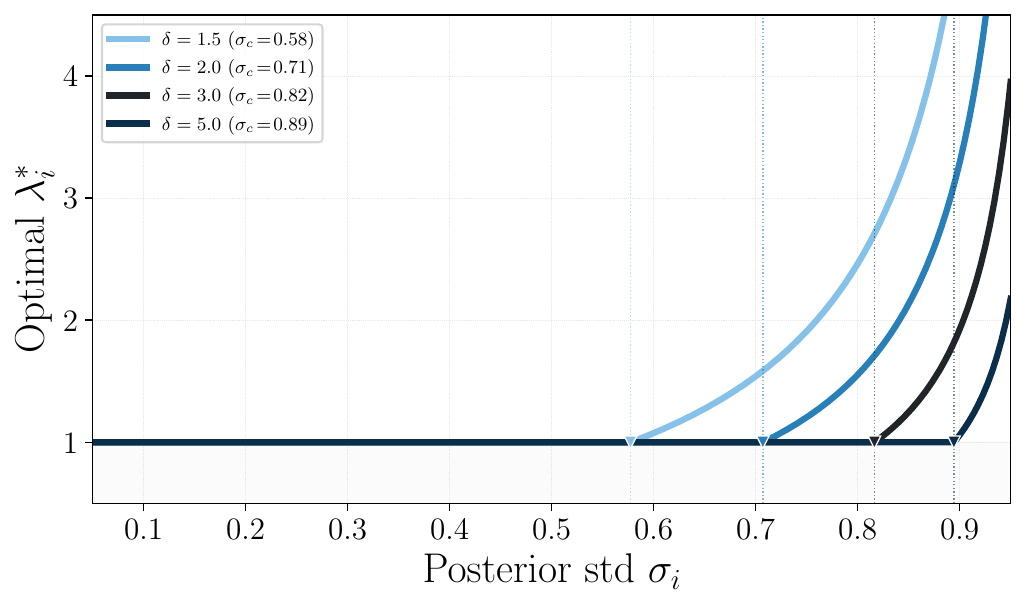}
  \caption{Optimal $\lambda^*_i(\sigma_i, \delta)$}
  \label{fig:optimal_lambda}
\end{subfigure}
\caption{\textbf{Variance equalization in $\lambda$-VAE.}
\emph{(a)} Convergence dynamics for $\lambda = 2$ ($\sigma^* = 0.5$): six
dimensions starting at $\sigma_0 \in \{0.10, 0.20, 0.35, 0.65, 0.80, 0.90\}$
all converge to $\sigma^* = 0.5$. Trajectories are solutions of
$d\sigma/dt = \sigma - \lambda\sigma^\lambda$ (the gradient correction term
with fixed reconstruction signal), isolating the equalization mechanism.
\emph{(b)} Optimal scaling exponent $\lambda^*_i(\sigma_i, \delta)$ for
$\delta \in \{2, 3, 5, 10\}$. Below $\sigma_c = \sqrt{1-1/\delta}$,
$\lambda^* = 1$; above it, $\lambda^*$ increases with $\sigma_i$, allocating
larger corrections to the most uncertain dimensions.}
\label{fig:equalization}
\end{figure}

Consider the gradient correction introduced by $\lambda$-scaling, except this
time with respect to $\log\sigma_i$; written as $ \nabla_i =
(\lambda\sigma_i^\lambda - \sigma_i)\cdot g_i^\sigma$. Setting the gradient
$\nabla_i = 0$, gives the equilibrium condition $\lambda\sigma_i^{\lambda-1}
= 1$ which is solved by
\begin{align}
    \sigma_i^* = \lambda^{-1/(\lambda-1)}.
    \label{eq:sigma_star}
\end{align}
Full derivation and stability proof in Appendix~\ref{app:sigma_star}.
This value is the same regardless of starting initial posterior variance. Hence,
every dimension, whether near-collapsed or over-sharp, is drawn toward the same
$\sigma^*$, provided they share the same $\lambda$. This is the content of
\emph{variance equalization}. In addition, the attractor defined
by Eq.~\ref{eq:sigma_star} is stable. That is, for $\sigma_i > \sigma^*$,
$\nabla_i < 0$ (system pushes $\sigma_i$ down); for $\sigma_i < \sigma^*$,
$\nabla_i > 0$ (system pushes $\sigma_i$ up). The correction is a bidirectional
stable restoring force. Furthermore, the linearization rate is $1 - \lambda$ for
$\lambda > 1$, so convergence is exponentially fast with rate $|\lambda - 1|$.
Figure~\ref{fig:equalization_convergence} illustrates the target
$\sigma^*(\lambda)$ and convergence from multiple starting points.

Hence, the polarized $\sigma$ distribution that characterizes emergent collapse
is corrected simultaneously across all dimensions. Similarly, the variance
equalization leads to emergent distributed representation as opposed to sparse
where entropy of the approximate posterior is
maximized~\cite{hinton1986learning,bengio2013representation}.

 
\subsection{Optimal $\lambda^*$ per Dimension}
\label{subsec:optimal}
 
Increasing $\lambda$ improves both gradient ratios and information gain but
risks divergence from the prior, which is critical for random generation. The
$\mathbb{KL}$ divergence from $\mathcal{N}(0, \mathbf{I})$ under the
$\lambda$-scaling is 
\begin{align}
    \mathbb{KL}(\lambda_i)
    = \frac{1}{2}\!\left(
      \sigma_i^{2\lambda_i} - \sigma_i^2
      - \log\sigma_i^{2\lambda_i} + \log\sigma_i^2
    \right) > 0
    \quad\text{for }\lambda_i > 1, \sigma_i < 1.
    \label{eq:mismatch}
\end{align}
As a result, we define an objective that can be used to calculate 
optimal $\lambda$ such that the information gain is balanced with the 
$\mathbb{KL}$ divergence, as follows

\begin{align}
    \mathcal{J}(\lambda_i, \sigma_i)
    = (\lambda_i - 1)|\!\log\sigma_i|
    - \delta\,\mathbb{KL}(\lambda_i), \qquad \delta > 1.
    \label{eq:objective}
\end{align}
The first term is the per-dimension gain from Eq.~\ref{eq:info_gain}; the second
term is used to penalize the additional $\mathbb{KL}$ divergence due to
$\lambda_i$; it is weighted by a hyperparameter $\delta > 1$ controlling the
tradeoff. The objective is strictly concave in $\lambda_i$ and has a unique
maximum, which is given by the following proposition. 
 
\begin{prop}[Optimal $\lambda^*$]
\label{prop:optimal_lambda}
The objective $\mathcal{J}$ is strictly concave in $\lambda_i$.
Its unconstrained maximum, subject to $\lambda_i \geq 1$, is:
\begin{align}
    \lambda^*_i
    = \max\!\left(1,\;\frac{\log(1 - 1/\delta)}{2\log\sigma_i}\right).
    \label{eq:lambda_star}
\end{align}
Full derivation in Appendix~\ref{app:lambda_star}.
\end{prop}
 
\noindent
The optimal solution provided by the above proposition has a
direct interpretation as \emph{optimal resource allocation}.
The estimate $\lambda^*_i \propto 1/|\log\sigma_i|$, hence uncertain 
dimensions (large $\sigma_i$, small $|\log\sigma_i|$) receive larger
corrections, while confident dimensions receive smaller ones or none at all.

Meanwhile, the hyperparameter $\delta$ plays a key role in determining the
aggressiveness of correction regime. Importantly, it determines a critical threshold
$\sigma_c$ as 
\begin{align}
\sigma_c := \sqrt{1-1/\delta}
\label{eq:critical_point}
\end{align}
For any $\sigma_i < \sigma_c$, $\lambda^{*}=1$, the $\mathbb{KL}$ divergence
outweighs the gains, hence the max in Eq.~\ref{eq:lambda_star} passes one. The
larger $\delta$, the less tolerant the system is of $\mathbb{KL}$ divergence. In
the extreme case, as $\delta \to \infty$ then $\lambda^*_i \to 1$ for all
dimensions, recovering the standard VAE. On the contrary, as $\delta \to 1^{+}$,
the system grows more tolerant to disparity in $\mathbb{KL}$; approaching a
deterministic autoencoder, see Figure~\ref{fig:optimal_lambda} for this
mechanism.
 
 
\section{Related Works}
\label{sec:related_works}
Posterior collapse in VAEs has motivated research along three lines: modifying
the $\mathbb{KL}$ term or constraining decoder capacity to address gradient imbalance;
isolating and minimizing marginal mismatch to improve aggregate posterior
alignment; and enriching the prior or posterior family. We discuss how each
relates to $\lambda$-VAE.

In $\beta$-VAE~\cite{higgins2016beta,burgess2018understanding} ELBO is modified
by weighting the $\mathbb{KL}$ term. Large $\beta > 1$ encourages
disentanglement but can encourage the decoder not to use latent information
early in training, leading to posterior collapse. In~\cite{fu2019cyclical},
cyclical annealing of $\beta$ is introduced between near zero and its maximum
value. In both cases, the training objective is modified globally, affecting
both reconstruction and $\mathbb{KL}$ gradients symmetrically, while offering no
per-dimension adaptivity. Free Bits~\cite{chen2016variational} imposes a
per-dimension minimum $\mathbb{KL}$ floor, introducing discontinuous gradients
but guaranteeing minimum information usage.
$\delta$-VAE~\cite{razavi2019preventing} enforces a similar constraint
differentiably, leaving the threshold per dimension to be tuned. None of the
above approaches introduce adaptive control over information flow conditioned on
local encoder confidence $\sigma_i$ as is done in $\lambda$-VAE. Regardless,
since $\lambda$-VAE modifies the sampling step rather than the training
objective, it is fully compatible with $\beta$-VAE, cyclical annealing, and
other $\mathbb{KL}$-weighting schedules and can be composed freely.

Normalising flows~\cite{rezende2015variational,kingma2016improved} enrich the
posterior family through invertible transformations, reducing the approximation
gap at the cost of significant parameter overhead. Rosca et al.~\cite{rosca2018distribution}
show that even expressive posteriors do not guarantee good aggregate matching,
consistent with our finding that the information gap and marginal mismatch can
be tracked to the same underlying issue.
 
Lagging inference~\cite{he2019lagging} updates the encoder more aggressively
than the decoder to prevent early collapse, targeting gradient imbalance but not
the information gap or marginal mismatch. Linear VAE
analyses~\cite{lucas2019don,ichikawa2024learning} formalize collapse as a phase
transition, while Ichikawa et al.~\cite{ichikawa2024learning} derive a closed-form
collapse threshold as a function of data covariance and $\mathbb{KL}$ weight.

 
Alternatively, using the ELBO decomposition~\cite{hoffman2016elbo},
InfoVAE~\cite{zhao2019infovae} introduces separate weights for $I_q(X;Z)$ and
the marginal, enforcing aggregate matching via maximum mean discrepancy (MMD).
Wasserstein autoencoders~\cite{tolstikhin2017wasserstein} match the aggregate
using optimal transport; adversarial
approaches~\cite{makhzani2015adversarial,mescheder2017adversarial} use
discriminator training. All require integration over the dataset to estimate
$q(z)$, adding compute cost and training complexity.
VampPrior~\cite{tomczak2018vae} learns a flexible prior as a mixture of
posteriors, reducing structural mismatch but requiring Monte Carlo $\mathbb{KL}$
estimation. Our duality result Eq.~\ref{eq:duality} shows that $\lambda$-VAE
addresses marginal mismatch implicitly through the information gap, without
aggregate estimation. Furthermore, we can explicity tradeoff
genaration-reconstruction quality using $\delta$ in Eq~\ref{eq:lambda_star}.

\section{Experiments}
\label{sec:experiments}
In this section we empirically validate $\lambda$-VAE on synthetic and real
datasets, demonstrating its effect on posterior collapse, variance equalization,
information flow, and generation quality.

\subsection{Dataset and Setup}
\label{sec:setup}
 
\noindent\textbf{Datasets:}
We use a combination of synthetic and collected datasets. The synthetic dataset
is drawn from a mixture of Gaussians $p(\mathbf{x}) = \sum_k \pi_k
\mathcal{N}(\mathbf{x};\,\mu_k, \sigma_k I)$, with $k{=}2$ and $k{=}4$
components, generating two-dimensional data. For binary tasks we use Binarized
MNIST~\cite{salakhutdinov2008quantitative,burda2015importance} and Binarized
OMNIGLOT~\cite{lake2015human,burda2015importance}; pixels are binarized
stochastically as described in~\cite{burda2015importance}. For RGB tasks we use
CIFAR-10~\cite{krizhevsky2009learning} ($32{\times}32$, 10 classes) and
CelebA-64~\cite{liu2015deep} (face images center-cropped and resized to
$64{\times}64$). All datasets use standard train–test splits and all results are
reported on the test set.

\noindent\textbf{Models:}
For synthetic and binarized datasets we use a 3-layer MLP with 300 hidden units
per layer for both encoder and decoder. For RGB datasets we use an
\emph{asymmetric} ResNet: a weak encoder (base channels 16, 1 residual block per
stage, 3 downsampling stages, channel multipliers $[1,2,4]$) paired with a
strong decoder (base channels 64, 2 residual blocks per stage, 3 upsampling
stages). The latent dimension is $K{=}2$ (synthetic), $K{=}30$ (MNIST and
OMNIGLOT), $K{=}512$ (CIFAR-10), and $K{=}1024$ (CelebA-64). We use Bernoulli
likelihood for binarized datasets and discretized logistic mixture (10
components) for RGB datasets. All likelihoods model pixels as statistically
independent except in the cases where PixelCNN
decoder~\cite{gulrajani2016pixelvae} is used-- we explicitly state it in such a
case.

\noindent\textbf{Training:}
Binary models are trained for 200 epochs with Adam~\cite{kingma2014adam}
(lr$=0.001$). Models on Natural Images are trained for 400 epochs on 8 GPUs
using Distributed Data Parallelism (DDP) with linear learning rate scaling
(effective lr$=8{\times}10^{-4}$)~\cite{goyal2017accurate}, cosine decay to
lr$_\text{min}=10^{-5}$, and gradient clipping at 1.0, with 100 batch size. For
optimal $\lambda^{*}$, the per-dimension values are updated every 5 epochs using
an exponential moving average (EMA) of $\sigma$ (decay $0.9$) and
Eq.~\ref{eq:lambda_star}; a linear ramp of 150 epochs is applied to prevent
premature compression before the encoder stabilizes. Meanwhile, the optimal
$\lambda$ in both binary and synthetic dataset is updated every 2 epochs. We set
$\delta{=}1.01$ for binary datasets and $\delta{=}1.001$ for image datasets,
unless we are reporting an ablation  study on the hyperparameter. The baseline is
an identical-architecture VAE with fixed $\lambda{=}1$ throughout training.
Hence, the online difference between optimal $\lambda$-VAE and Standard VAE, is
the per-dimension $\lambda$ update according to Eq.~\ref{eq:lambda_star}.

\noindent\textbf{Metrics:}
We report NLL (nats) and bits per dimension $\text{BPD} = \text{NLL}\,/\,(D\ln
2)$ computed as the ELBO on the full test set\footnote{%
  IWAE~\cite{burda2015importance} is systematically biased for $\lambda$-VAE
  models: when $\sigma^{\lambda_d} \ll 1$ all importance samples cluster near
  $\mu$, collapsing the effective sample size and overestimating NLL\@. Hence,
  we report ELBO-BPD as the primary metric.}.
Active units (AU) count dimensions with $A_{z_i} =
\mathrm{Cov}_x\!\left(\mathbb{E}[z_i|x]\right) >
0.01$~\cite{burda2015importance}.
For binary and image datasets we estimate $I(X;Z)$ via the SNR-based
\emph{information capacity} derived directly from the Gaussian channel model of
Section~\ref{subsec:info_gain} (Eq.~\ref{eq:gaussian_channel}).
For synthetic data the true $I(X;Z)$ is bounded by the $K{=}2$ latent
space and remains well below $\log N$ for any practical $N$; we therefore
use the Monte Carlo mixture marginalisation estimator
$\hat{I} = H(Z) - H(Z|X)$, where $H(Z)$ is approximated by the
$N$-component posterior mixture and $H(Z|X)$ is evaluated in closed form.
We report $C$ as \textbf{Capacity} in
Tables~\ref{tab:mnist_omniglot} and~\ref{tab:image_datasets}, and
the information gap $I(X;\phi) - C$ to track how much encoder capacity
reaches the decoder.
We additionally report \textbf{decoder capacity} $\sum_i C_i^{\mathrm{dec}}$,
where
\begin{align}
  C_i^{\mathrm{dec}}
  \;=\; \sqrt{\,\mathbb{E}_x\!\left[\!\left(
        \frac{\partial \log p(x|z)}{\partial z_i}
        \right)^{\!2}\right]}
  \label{eq:decoder_capacity}
\end{align}
is the RMS decoder gradient per latent dimension
(Tables~\ref{tab:image_datasets} and~\ref{tab:auto_regression}); a dimension
with $C_i^{\mathrm{dec}} \approx 0$ contributes nothing to the reconstruction
regardless of $\sigma_i$.

\begin{figure}[t]
\centering
\begin{subfigure}[t]{0.47\linewidth}
  \centering
  \includegraphics[width=\linewidth]{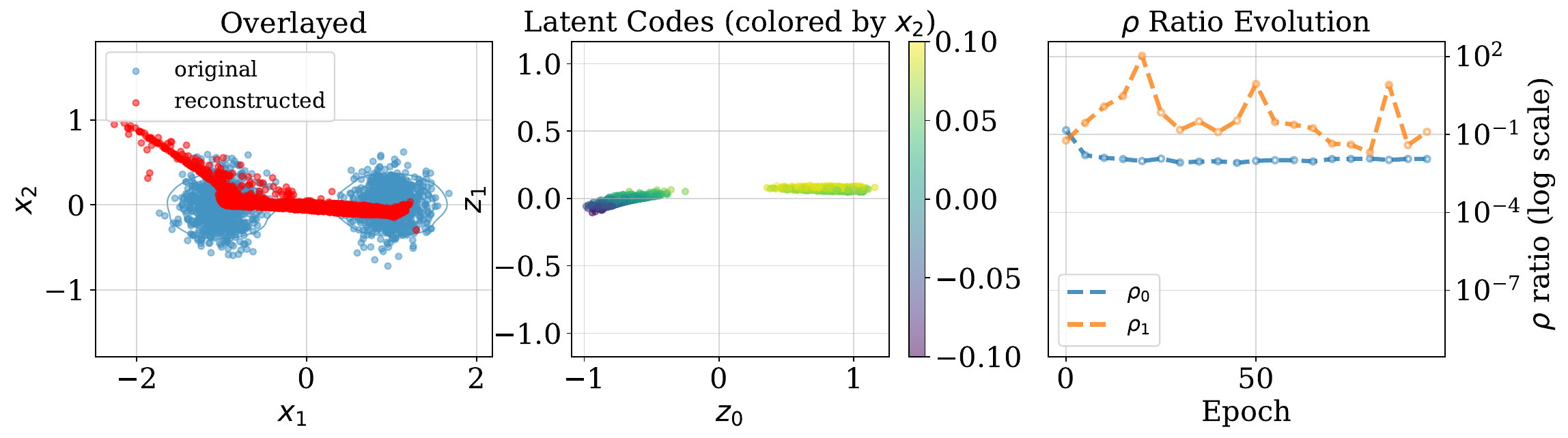}
  \caption{Gradient imbalance ($k{=}2$): standard VAE.}
  \label{fig:vanilla_grad_imbalance}
\end{subfigure}
\hspace{0.04\linewidth}
\begin{subfigure}[t]{0.47\linewidth}
  \centering
  \includegraphics[width=\linewidth]{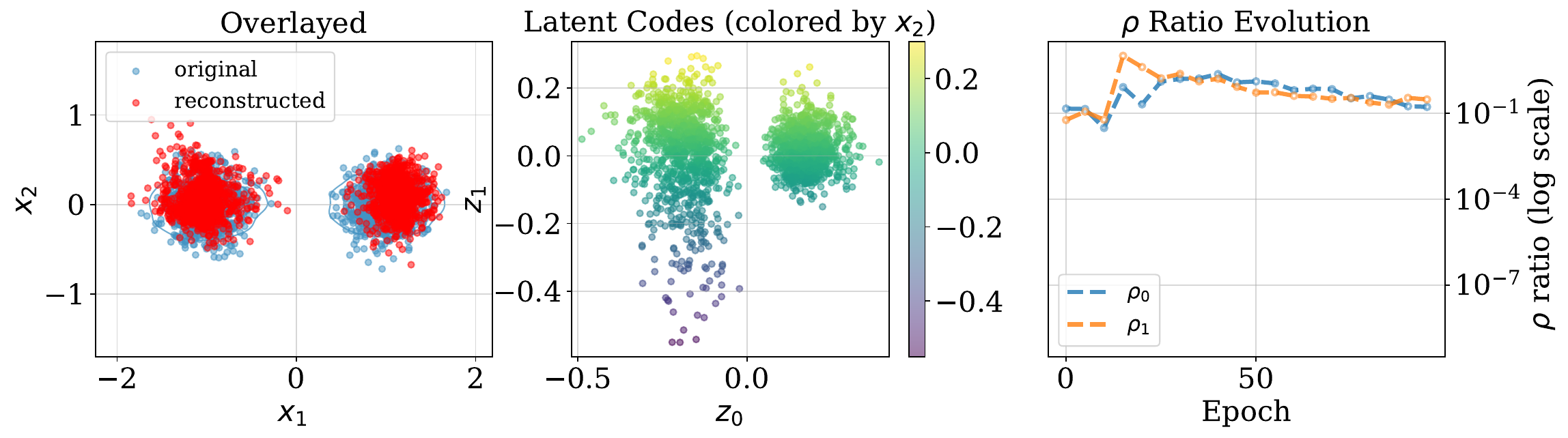}
  \caption{Gradient imbalance ($k{=}2$): $\lambda$-VAE ($\delta{=}1.001$).}
  \label{fig:lambda_grad_imbalance}
\end{subfigure}
\\[6pt]
\begin{subfigure}[t]{0.47\linewidth}
  \centering
  \includegraphics[width=\linewidth]{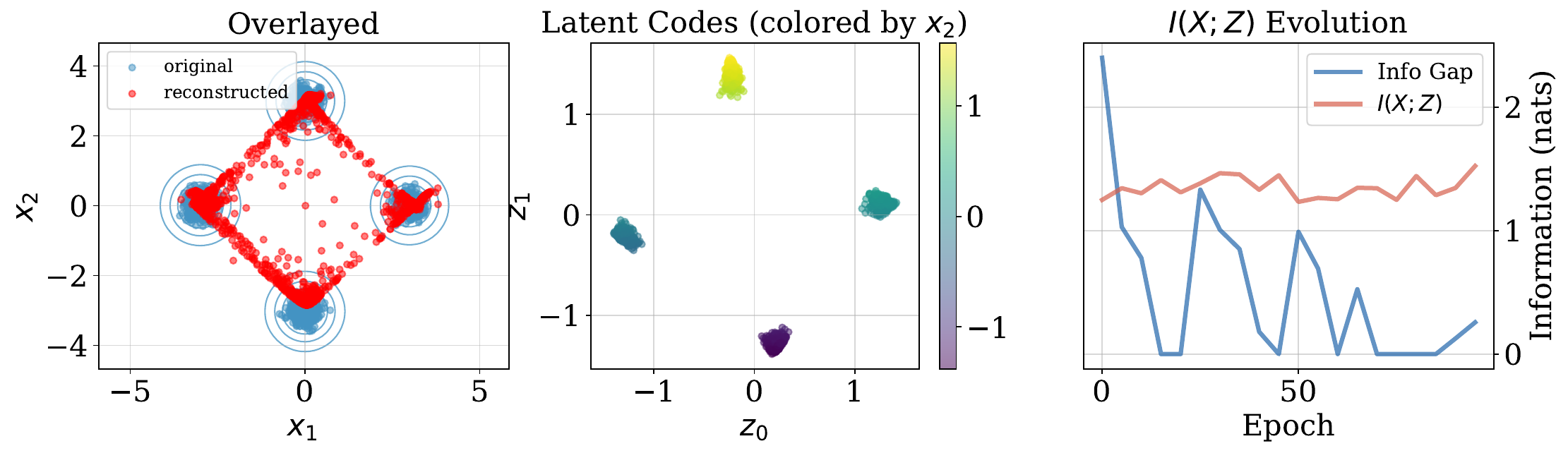}
  \caption{Information gap ($k{=}4$): standard VAE.}
  \label{fig:vanilla_info_gap}
\end{subfigure}
\hspace{0.04\linewidth}
\begin{subfigure}[t]{0.47\linewidth}
  \centering
  \includegraphics[width=\linewidth]{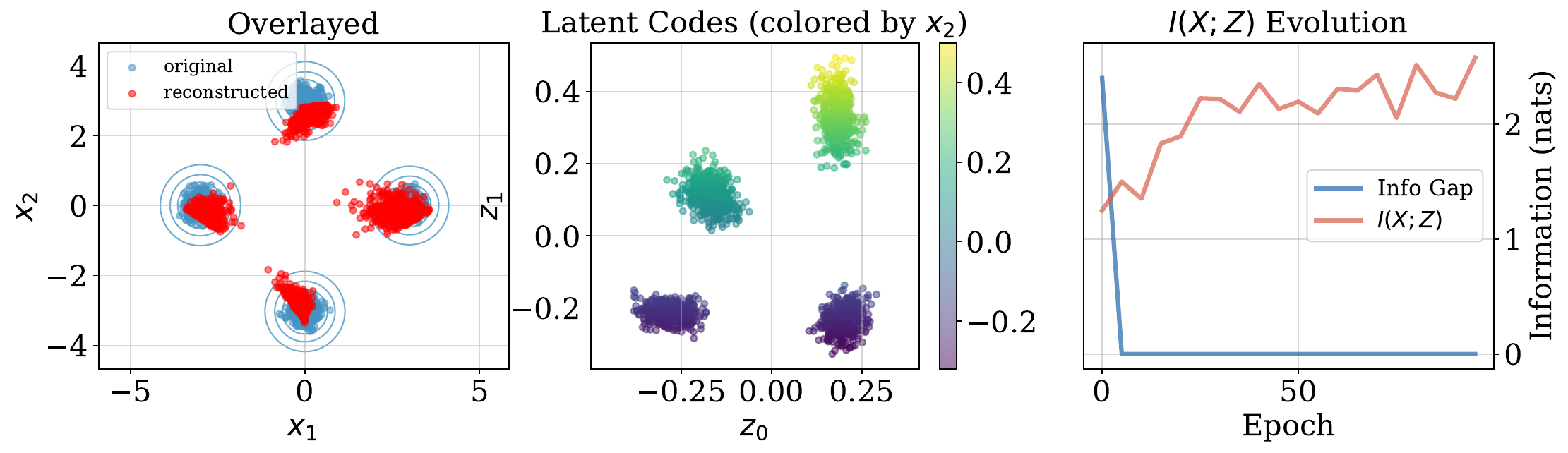}
  \caption{Information gap ($k{=}4$): $\lambda$-VAE ($\delta{=}1.001$).}
  \label{fig:lambda_info_gap}
\end{subfigure}
\caption{\textbf{Gradient imbalance and information gap on synthetic Gaussian
mixtures.} Each panel tracks the
gradient ratio $\rho_i$ and mutual information $I(X;Z)$ over training, alongside
the latent space. \emph{Top row} ($k{=}2$, clusters stretched along one axis),
isolating gradient imbalance. In the standard VAE the reconstruction gradient
$g_i^\sigma$ collapses to zero for the low-variance dimension before $\sigma_i
\to 1$, confirming Proposition~\ref{prop:ratio_grad}; the latent space shows
complete information loss as well. $\lambda$-VAE amplifies the gradient by
$\lambda\sigma_i^{\lambda-1}$, keeping $\rho_i$ elevated and the dimension
active throughout training. \emph{Bottom row} ($k{=}4$, equal inter-cluster
separation along each axis): the standard VAE exhibits a fluctuating non-zero
information gap, resulting in variation loss in both the latent and
reconstruction spaces. Under $\lambda$-VAE with $\lambda^*{>}2$ for all
dimensions the gap is driven to zero and the four-cluster structure is
recovered, consistent with Section~\ref{subsec:info_gain}.}
\label{fig:latent_collapse}
\end{figure}

\begin{table}[t]
\caption{\textbf{Comparison on binarized benchmarks (MNIST and Omniglot, $K{=}30$).} $\lambda$-VAE and the ablations shown ($\beta$-VAE,
standard VAE) use a three-layer MLP encoder--decoder; most published baselines
use deeper convolutional architectures, making these comparisons indicative
rather than directly controlled. VampPrior is an exception, using a comparable
MLP architecture. NLL and BPD measure reconstruction quality; AU counts active
latent dimensions; Capacity is the SNR information capacity, measuring how much
information reaches the decoder. Results for external baselines are taken from
original papers.}
\label{tab:mnist_omniglot}
\centering
\small
\setlength{\tabcolsep}{5pt}
\begin{tabular}{llcccc}
\toprule
\textbf{Dataset} & \textbf{Model}
& \textbf{NLL}$\downarrow$ & \textbf{BPD}$\downarrow$
& \textbf{AU}$\uparrow$  & \textbf{Capacity (nats)}$\uparrow$ \\
\midrule
\multirow{10}{*}{\textbf{MNIST}}
 & Free Bits~\cite{kingma2016improved}          & 79.10 & 0.145  & -- & -- \\
 & AVB~\cite{mescheder2017adversarial}          & 80.24 & 0.147 & -- & -- \\
 & VampPrior~\cite{tomczak2018vae}              & 85.57 & 0.157 & -- & -- \\
 & InfoVAE~\cite{zhao2019infovae}               & 80.76 & 0.148 & -- & -- \\
 & $\beta$-VAE ($\beta\!=\!0.5$)~\cite{higgins2016beta} & 102.94 & 0.189 & 18 / 30 & 34.2 \\
 & $\beta$-VAE ($\beta\!=\!2.0$)                & 96.73  & 0.178 & 11 / 30 & 16.2 \\
 & Standard VAE ($\lambda=1.0$)                  & 97.17  & 0.178 & 14 / 30 &
 23.2 \\
  & $\bm{\lambda}$-\textbf{VAE} ($\delta\!=\!1.19$) & 93.80 & 0.172 & 21 / 30 & 29.5 \\
 & $\bm{\lambda}$-\textbf{VAE} ($\delta\!=\!1.10$) & 94.44 & 0.173 & 23 / 30 & 37.2 \\
 & $\bm{\lambda}$-\textbf{VAE} ($\delta\!=\!1.01$) & \textbf{78.43} & \textbf{0.144} & \textbf{29 / 30} & \textbf{69.4} \\
\midrule
\multirow{8}{*}{\textbf{OMNIGLOT}}
 & VampPrior                                    & 104.75 & 0.192 & -- & -- \\
 & $\beta$-VAE ($\beta\!=\!0.5$)                & 131.65 & 0.242 & 21 / 30 & 38.4 \\
 & $\beta$-VAE ($\beta\!=\!2.0$)                & 119.86 & 0.220 & 13 / 30 & 15.0 \\
 & Standard VAE ($\lambda=1.0$)                  & 123.11 & 0.226 & 17 / 30 &
 25.6 \\
 & $\bm{\lambda}$-\textbf{VAE} ($\delta\!=\!1.19$) & 111.14 & 0.204 & 29 / 30 & 32.4 \\
 & $\bm{\lambda}$-\textbf{VAE} ($\delta\!=\!1.10$) & 110.54 & 0.203 & 30 / 30 & 37.4 \\
 & $\bm{\lambda}$-\textbf{VAE} ($\delta\!=\!1.01$) & \textbf{103.03} & \textbf{0.189} & \textbf{30 / 30} & \textbf{67.7} \\
\bottomrule
\end{tabular}
\end{table}

\begin{figure}[t]
\centering
\begin{subfigure}[t]{0.5\linewidth}
  \centering
  \includegraphics[width=\linewidth]{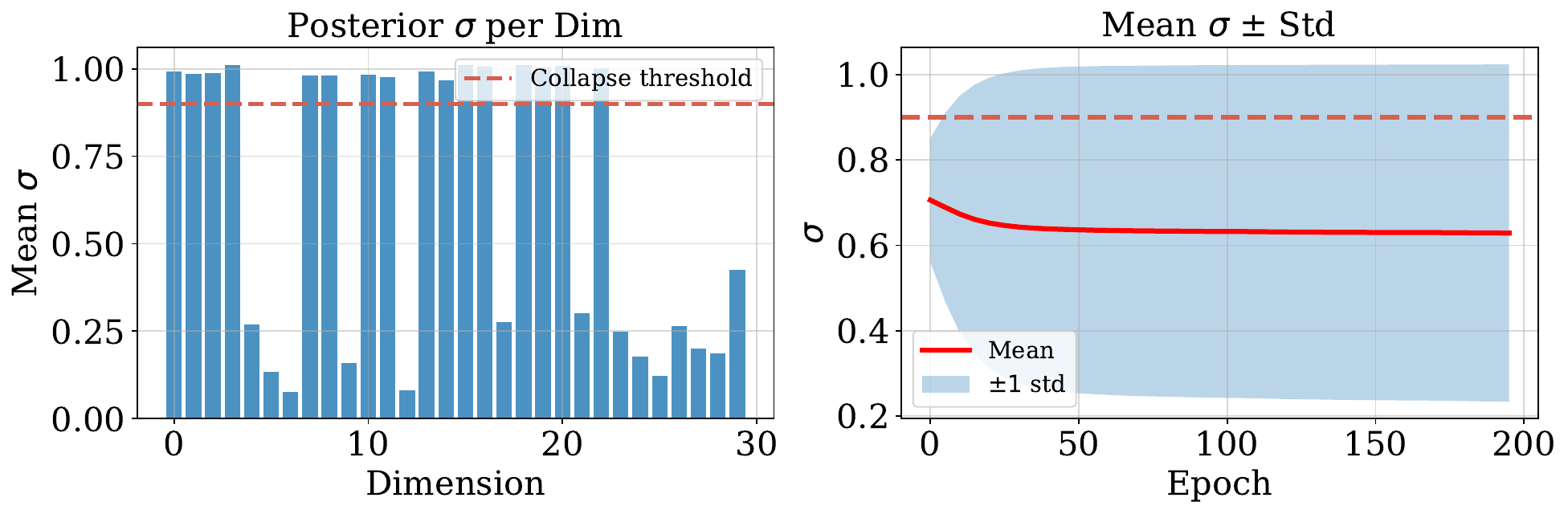}
  \caption{Standard VAE: $\sigma$ histogram and trajectory.}
  \label{fig:vanilla_eq}
\end{subfigure}\hfill
\begin{subfigure}[t]{0.5\linewidth}
  \centering
  \includegraphics[width=\linewidth]{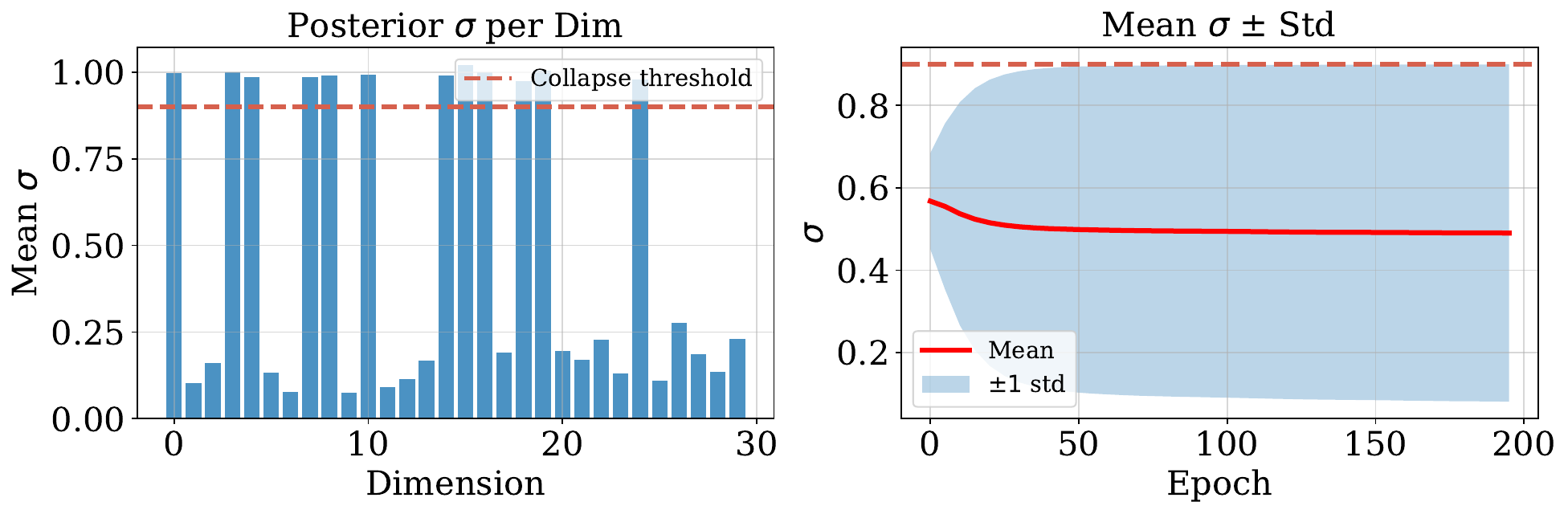}
  \caption{$\beta$-VAE ($\beta{=}0.5$): $\sigma$ histogram and trajectory.}
  \label{fig:beta_eq}
\end{subfigure}
\\
\begin{subfigure}[t]{0.5\linewidth}
  \centering
  \includegraphics[width=\linewidth]{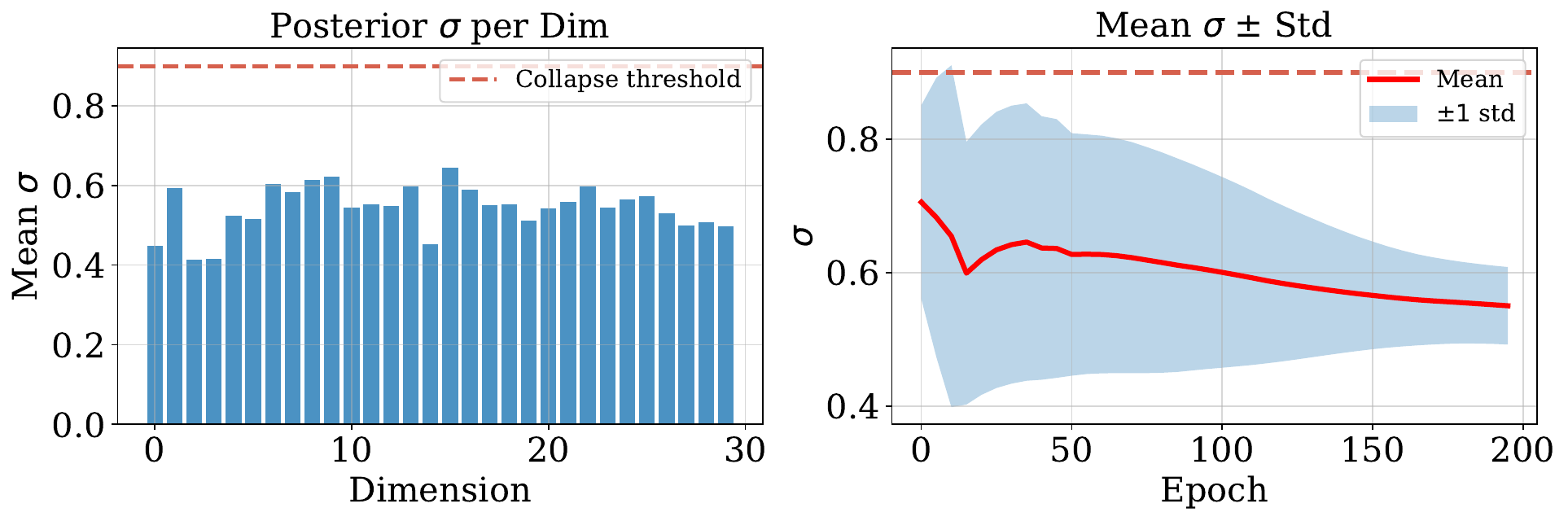}
  \caption{$\lambda$-VAE ($\delta=1.001$): variance equalization.}
  \label{fig:optimal_eq}
\end{subfigure}\hfill
\begin{subfigure}[t]{0.5\linewidth}
  \centering
  \includegraphics[width=\linewidth]{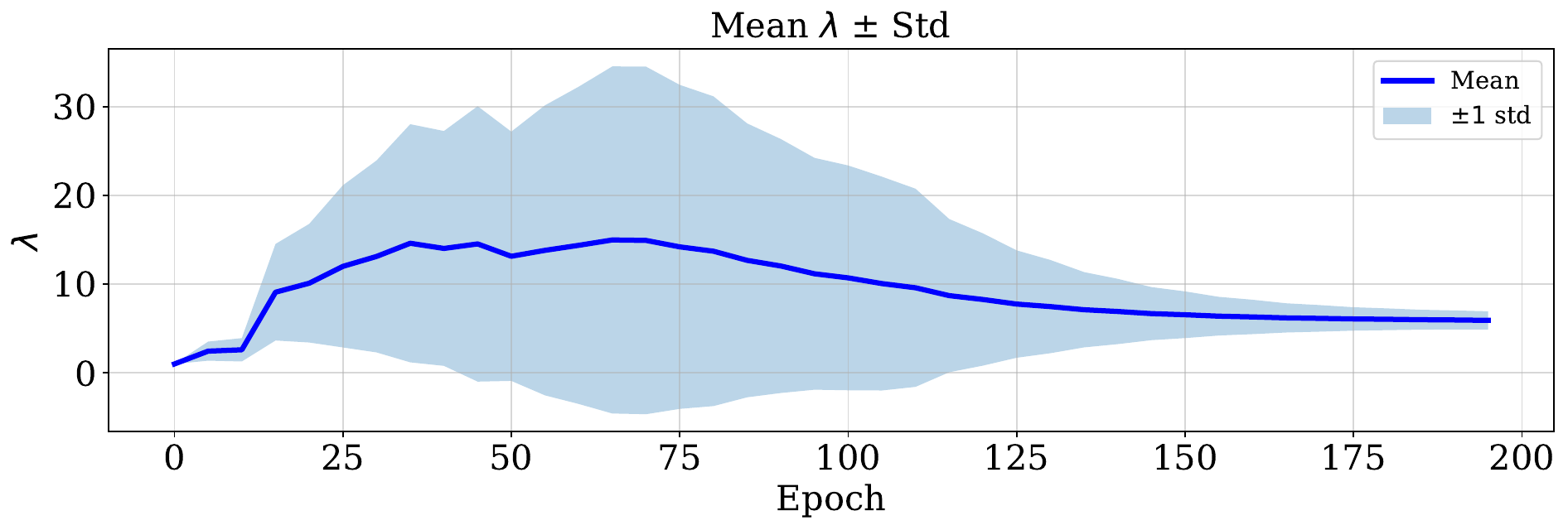}
  \caption{$\bar{\lambda}^{*}$ trajectory (mean $\pm$ std over dimensions)}
  \label{fig:lambda_traj}
\end{subfigure}
\caption{\textbf{Variance equalisation on Binary-MNIST ($K{=}30$).}
Panels~(a)--(c) each show the per-dimension $\sigma$ histogram alongside the
mean $\pm$ std trajectory over training epochs.
\emph{(a)} Standard VAE: the distribution becomes bimodal; collapsed dimensions
accumulate near $\sigma{=}1$ while active dimensions cluster at lower values.
\emph{(b)} $\beta$-VAE ($\beta{=}0.5$): the mode shifts downward but a
tail near $\sigma{=}1$ persists.
\emph{(c)} $\lambda$-VAE ($\delta{=}1.001$): all dimensions converge to a
tight band around $\sigma^*$; the across-dimension std collapses to near
zero, realising variance equalization (Section~\ref{subsec:equalization}).
\emph{(d)} Corresponding $\bar\lambda$ trajectory (mean $\pm$ std over
dimensions): $\bar\lambda$ stabilises at ${\approx}\,5.3$, confirming joint
$\sigma^*$--$\lambda^*$ convergence.}
\label{fig:equalization_exp}
\end{figure}

\subsection{Collapse Prevention, Information Recovery, and Reconstruction Quality}
\label{sec:exp_posterior}

We validate $\lambda$-VAE along three axes, suppression of gradient imbalance,
reduction of the information gap, and downstream reconstruction quality.

\noindent\textbf{Synthetic data:} We ran a standard VAE and $\lambda$-VAE on two
synthetic $k$-component Gaussian mixtures with $K{=}2$ latent dimensions and
$\delta{=}1.001$, designed to simulate the two causes of collapse independently
(Figure~\ref{fig:latent_collapse}). For $k{=}2$, the reconstruction gradient
norm of the standard VAE's first dimension collapses to zero before $\sigma_1
\to 1$, matching Proposition~\ref{prop:ratio_grad}; the latent space and
reconstructions confirm complete information loss from that dimension.
$\lambda$-VAE amplifies the gradient by $\lambda\sigma^{\lambda-1}$, keeping the
dimension active throughout training; compare
Figure~\ref{fig:vanilla_grad_imbalance} vs ~\ref{fig:lambda_grad_imbalance}. For
$k{=}4$, the standard VAE exhibits a fluctuating yet non-zero information gap
that suppresses variation in both latent and data clusters, while introducing
high reconstruction error per dimension leading to noisy data recovery. Under
$\lambda$-VAE, however, since $\lambda^* > 2$ for all dimensions the gap is
driven to zero, consistent with Section~\ref{subsec:info_gain}, while both
latent and reconstruction spaces preserve the four-cluster structure relatively
well.

\noindent\textbf{Binary data:} On Binarized MNIST and Binarized OMNIGLOT
($K{=}30$, three-layer MLP), the standard VAE collapses 16 of 30 dimensions on
MNIST (AU$\,{=}\,14$) and 13 of 30 on OMNIGLOT (AU$\,{=}\,17$). $\lambda$-VAE
($\delta{=}1.01$) reduces this to one on MNIST (AU$\,{=}\,29/30$) and zero on
OMNIGLOT (AU$\,{=}\,30/30$), as reported in Table~\ref{tab:mnist_omniglot}. In
Figure~\ref{fig:equalization_exp}, the mechanism is illustrated via $\sigma$
distributions. The standard VAE is bimodal with a mass near $\sigma{=}1$, while
$\beta$-VAE shifts the distribution but does not eliminate the tail. The
$\lambda$-VAE, however, concentrates all mass in a narrow band around
$\sigma \approx 0.4$, realizing the variance equalization of
Section~\ref{subsec:equalization}. The average per-dimension estimate stabilizes
at $\bar\lambda \approx 5.3$ on MNIST and $\bar\lambda \approx 2.5$ on OMNIGLOT
at $\delta{=}1.01$, reflecting their different active-dimension $\sigma$
distributions; Section~\ref{sec:exp_equalization} confirms the theoretical
predictive curve Eq.~\ref{eq:lambda_star} aligns with empirical data.

The Capacity column in Table~\ref{tab:mnist_omniglot} separates two mechanisms
for increasing information throughput, computed as Eq.~\ref{eq:gaussian_channel}.
$\beta$-VAE ($\beta{=}0.5$) achieves comparable capacity to $\lambda$-VAE
($\delta{=}1.1$) on both datasets (34.2 vs.\ 37.2 nats on MNIST; 38.4 vs.\ 37.4
on OMNIGLOT), but does so by loosening the $\mathbb{KL}$ coefficient, inflating
$\mathrm{Var}_x[\mu_d]$ across active dimensions at the expense of prior
matching. The consequence is substantially worse BPD (0.242 vs.\ 0.203 on
OMNIGLOT), while the aggregate posterior drifts from $\mathcal{N}(0,\mathbf{I})$.
Provided a calibrated $\delta$ is chosen, $\lambda$-VAE instead reduces the
effective noise $\sigma_d^{\lambda_d}$ per dimension without weakening the
prior. At $\delta{=}1.01$, stronger per-dimension compression (avg.\
$\bar\lambda{\approx}5.3$ on MNIST) drives Capacity to 69.4 nats (nearly
$3\times$ the standard VAE), while simultaneously achieving the best BPD.
Meanwhile, $\delta=1.19$ achieves a comparable AU, despite a significantly lower
information throughput -- see Section~\ref{sec:exp_equalization} for discussion
on effects of $\delta$.

\begin{figure}[t]
  \centering
  \begin{subfigure}[t]{0.5\linewidth}
  \centering
  \includegraphics[width=\linewidth]{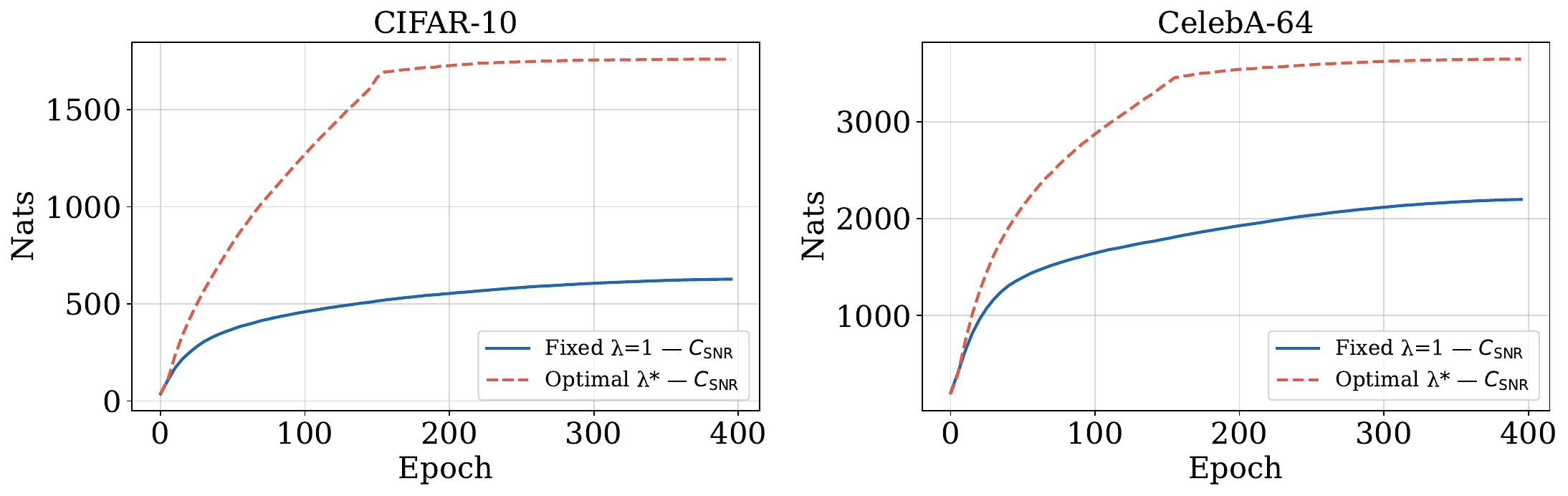}
  \caption{SNR information capacity $C_{\mathrm{SNR}}$}
  \label{fig:info_gap}
\end{subfigure}\hfill
\begin{subfigure}[t]{0.5\linewidth}
  \centering
  \includegraphics[width=\linewidth]{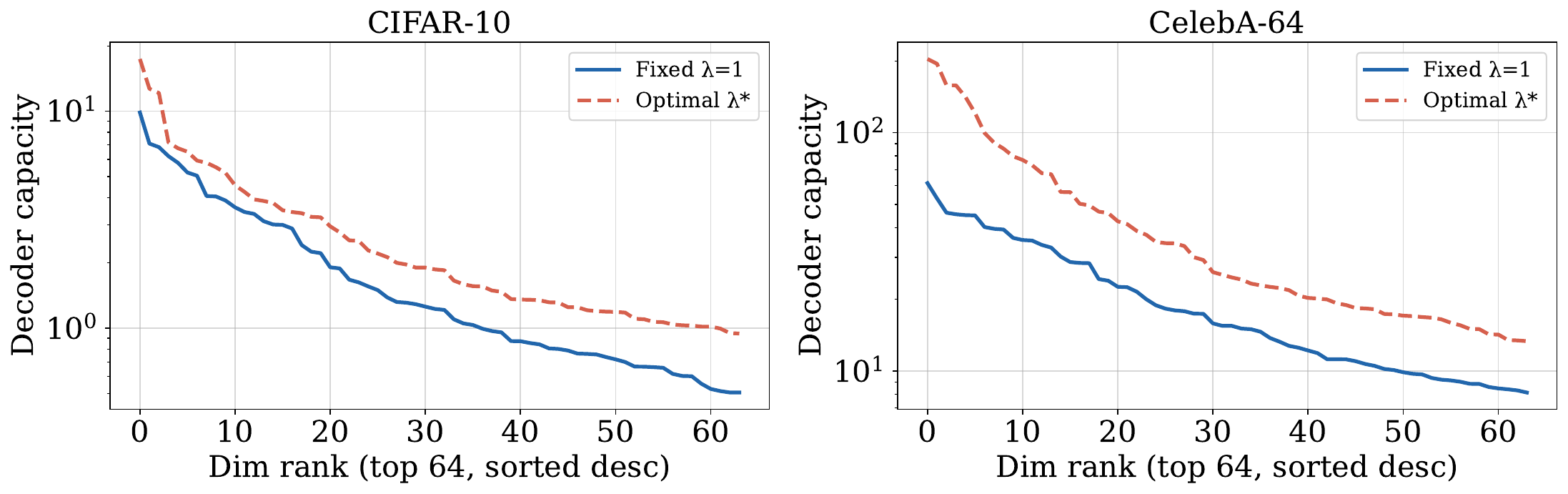}
  \caption{Decoder capacity.}
  \label{fig:decoder_capacity}
\end{subfigure}
  \caption{\textbf{SNR information capacity and decoder capacity over training.}
  (a) Under optimal $\lambda$, $C_{\mathrm{SNR}}$ grows from 628 to 1760 nats on
  CIFAR-10 ($2.8\times$) and from 2198 to 3646 nats on CelebA-64 ($1.7\times$).
  (b) Decoder capacity rises in parallel: 169$\to$361 nats ($2.1\times$) on
  CIFAR-10 and 3283$\to$6596 nats ($2.0\times$) on CelebA-64, confirming that
  the increased encoder throughput is actively utilised by the decoder.}
  \label{fig:info_gap_dynamics}
\end{figure}

\noindent\textbf{Natural images:} On RGB images, the fixed-$\lambda$ model on
CIFAR-10 leaves 173 of 512 dimensions (AU=339), i.e., dimensions where the
optimal schedule would assign $\lambda^*_d \gg 1$ but fixed $\lambda{=}1$
provides no correction.  The SNR information capacity
Eq.~\ref{eq:gaussian_channel} (Table~\ref{tab:image_datasets}) sits at 628 nats.
On the contrary under optimal $\lambda$, all 512 dimensions are driven below
$\sigma{=}0.9$ and the effective noise $\sigma_d^{\lambda_d}$ is equalized
across the full latent code; $C_{\mathrm{SNR}}$ grows to 1760 nats ($2.8\times$
improvement), as shown in Figure~\ref{fig:info_gap}.  Decoder capacity rises in
parallel from 169 to 361 nats ($2.1\times$, Figure~\ref{fig:decoder_capacity})
as a consequence of gradient rebalancing.  

On CelebA-64 ($K{=}1024$),
$C_{\mathrm{SNR}}$ grows from 2198 to 3646 nats ($1.7\times$) and decoder
capacity from 3283 to 6596 nats ($2.0\times$). These gains translate directly
into reconstruction quality. ELBO-BPD improves by $+0.33$ on CIFAR-10 (7.1\%
relative) and $+0.24$ on CelebA-64 (6.2\% relative), see
Table~\ref{tab:image_datasets}.

\begin{figure}[t]
  \centering
  \begin{subfigure}[t]{0.65\linewidth}
  \centering
  \includegraphics[width=\linewidth]{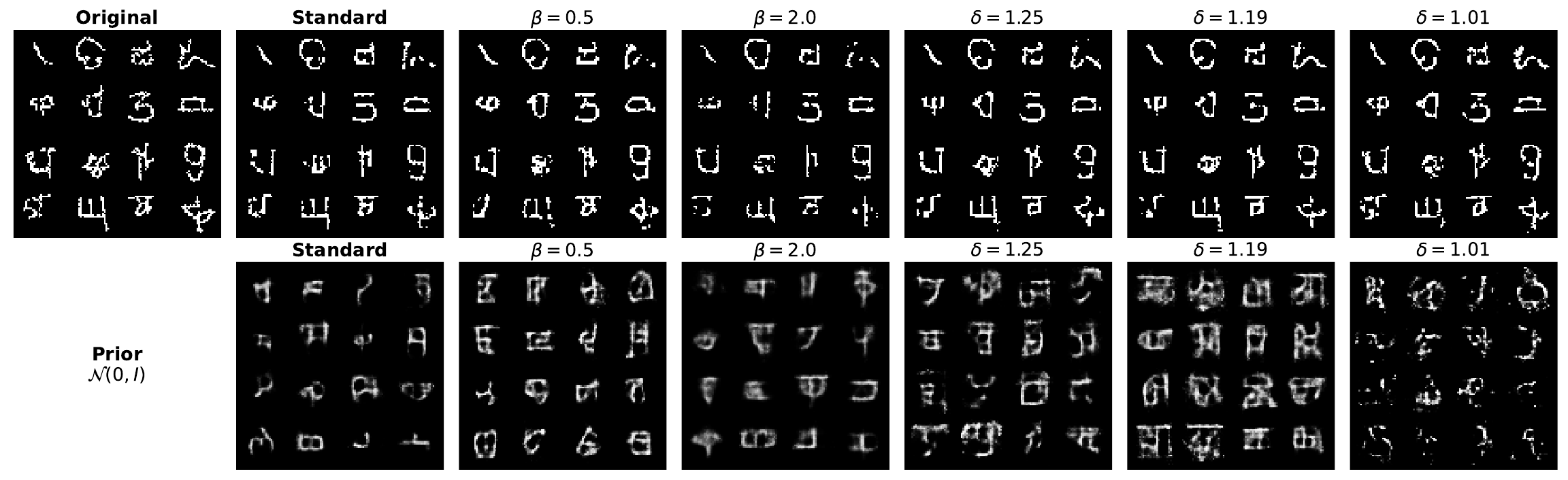}
  \caption{Reconstruction vs Random sampling}
  \label{fig:Recon_vs_generation}
\end{subfigure}\hfill
\begin{subfigure}[t]{0.34\linewidth}
  \centering
  \includegraphics[width=\linewidth]{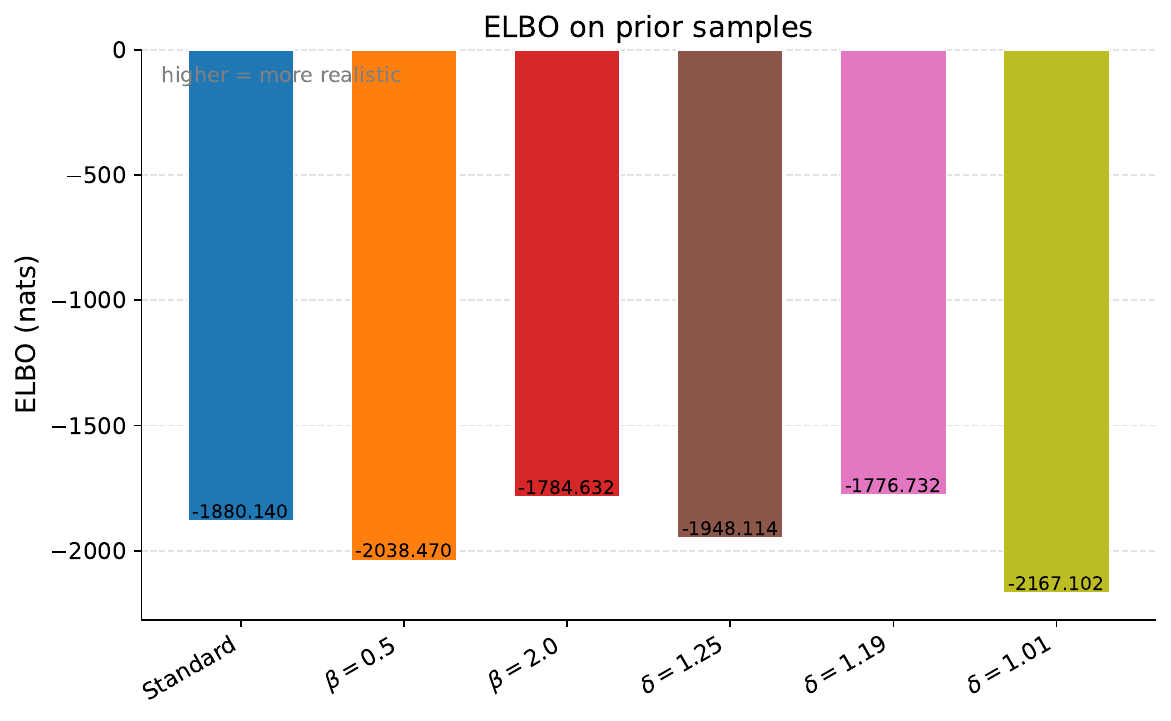}
  \caption{Prior samples ELBO}
  \label{fig:ELBO_sample}
\end{subfigure}
  \caption{\textbf{Reconstruction vs Random sampling.}
    Reconstruction and prior sampling on Omniglot; figures best viewed zoomed.
    (a) Top: test reconstructions ($z{=}\mu$); bottom: samples from
    $z\sim\mathcal{N}(0,I)$ displayed as decoder probabilities as pixel intensity
    for prob > 0.5 (greedy sampling of Bernoulli). Random samples are shared
    across models for the decoding. (b) ELBO on 256 prior samples per model
    variant. The figures show that $\delta{=}1.19$ retains reconstruction
    fidelity while producing coherent prior samples; $\delta{=}1.01$ achieves
    sharp reconstructions but loses generative quality as the aggregate
    posterior drifts from $\mathcal{N}(0,I)$, confirming that $\delta$ can be
    calibrated to close the information gap without over-compressing the latent
    code.}
  \label{fig:sample_recons_abalation}
\end{figure}

\subsection{Optimal $\lambda$ and Its Sensitivity to $\delta$}
\label{sec:exp_equalization}

\begin{figure}[t]
  \centering
  \includegraphics[width=\linewidth]{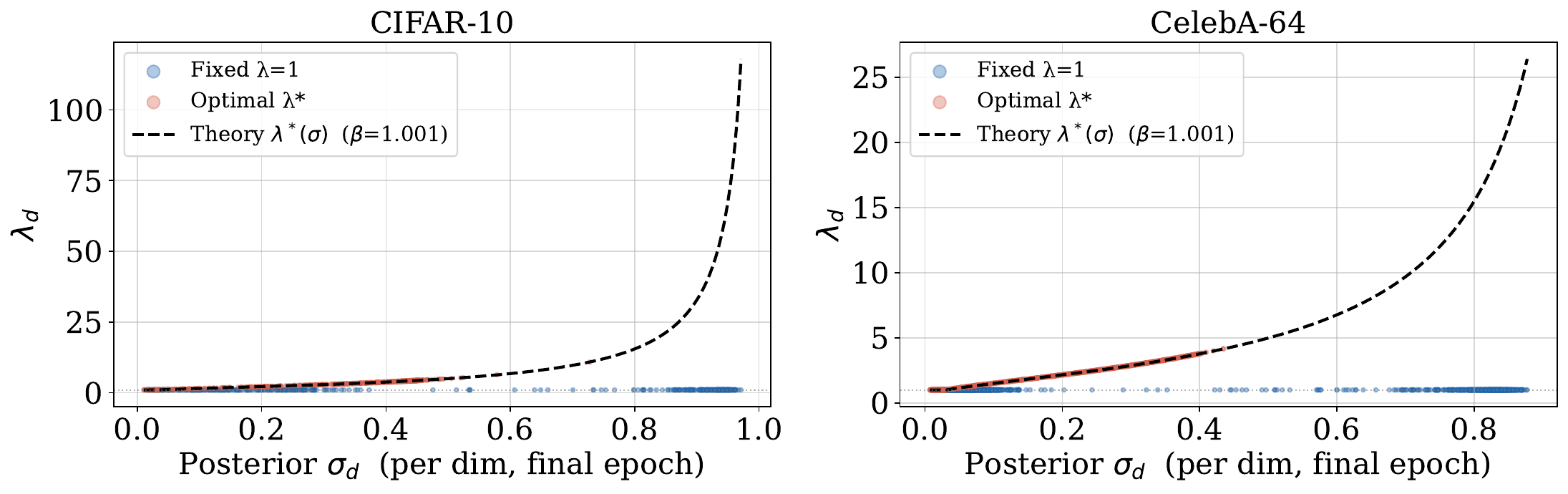}
  
  \caption{\textbf{Empirical alignment with the optimal $\lambda^*$ schedule.}
    All 512 CIFAR-10 and 1024 CelebA-64 dimensions lie exactly on the
    theoretical curve Eq.~\ref{eq:lambda_star} at convergence (Pearson $r = 1.00$). Points near $\sigma
    \approx 0$ have $\lambda^* \to \lambda_{\max} = 20$ (clamped); points near
    $\sigma = 1$ are dimensions the vanilla VAE model leaves dead.}
  \label{fig:lambda_scatter}
\end{figure}

\noindent\textbf{$\lambda^*$ estimate alignment with theory:}
Figure~\ref{fig:lambda_scatter} plots the per-dimension $(\sigma, \lambda^*)$
scatter for all latent dimensions in CIFAR-10  and CelebA-64 at epoch 400.  Every
dimension lies exactly on the theoretical curve defined in
Proposition~\ref{prop:optimal_lambda},  
confirming that the EMA-based update converges to the analytic optimum derived
in Appendix~\ref{app:optimal_lambda_derivation}.  The fixed model's dimensions
form a horizontal line at $\lambda{=}1$ across the full $\sigma$ range; the 173
dimensions with $\sigma > 0.9$ would receive $\lambda^* \gg 1$ under the optimal
schedule, representing a systematic under-utilisation that fixed $\lambda$
cannot correct.  The same alignment holds for CelebA-64 ($K{=}1024$;
$\bar\lambda{=}1.99$, $\lambda_{\max}{=}10.4$ at convergence). The convergence
of $\lambda^{*}$ and $\sigma^{*}$ is also shown in
Figure~\ref{fig:equalization_exp} for Binary MNIST dataset.

\noindent\textbf{Effect of $\delta$:} The parameter $\delta$ drives the
equilibrium $\sigma^*$ via optimal estimate Eq.~\ref{eq:lambda_star}. It controls
the aggressiveness of compression (how close sample $z$ is to $\mu$). Smaller
$\delta$ induces larger $\lambda^*$, compressing more dimensions more tightly.
Table~\ref{tab:mnist_omniglot} illustrates this particularly on binary MNIST
dataset; reducing $\delta$ from 1.19 to 1.01 increases AU from 21/30 to 29/30
and improves ELBO-BPD from 0.172 to 0.144. The AU ceiling, however, is reached at
$\delta{=}1.1$ for OMNIGLOT.

As discussed in Section~\ref{subsec:optimal}, the tradeoff here is increased
marginal divergence for improved information flow, which consequently affects
the quality of sampling from a prior. This can be calibrated as required.
Figure~\ref{fig:sample_recons_abalation} shows that $\delta{=}1.19$ navigates
this tradeoff effectively -- it improves reconstruction, BPD 0.204 vs.\ 0.226
for the standard VAE on OMNIGLOT, while achieving \emph{better} prior sample
quality than all compared methods (Standard VAE, $\beta$-VAE and different
$\delta$ values) with $-1776$ ELBO, estimated on 256 prior samples. Meanwhile,
$\delta{=}1.01$, while minimising BPD (0.189), substantially degrades generative
quality ($-2167$ ELBO). Consequently, choosing the right $\delta$ leads
to maximized net information gain Eq.~\ref{eq:objective}, with little or no
compromise on the generative quality.


\begin{table}[t]
\caption{\textbf{Controlled comparison on CIFAR-10 and CelebA-64.} Fixed and
optimal $\lambda$ share the same asymmetric ResNet architecture. SNR capacity
information throughput; decoder capacity (Figure~\ref{fig:decoder_capacity})
measures the information each latent dimension actively contributes to the
decoder output.}
\label{tab:image_datasets}
\centering
\small
\setlength{\tabcolsep}{5pt}
\begin{tabular}{llrrrrc}
\toprule
\textbf{Dataset} & \textbf{Model} &
  \textbf{BPD $\downarrow$} & \textbf{SNR cap.\ $\uparrow$} &
  \textbf{Dec.\ cap.\ $\uparrow$} & $\bar\lambda$ & \textbf{AU $\uparrow$} \\
\midrule
{\bf CIFAR-10} & Vanilla VAE                                    & 4.645 & 628  &
  169  & 1.00 & 339\,/\,512 \\
  & $\bm{\lambda}$-\textbf{VAE} ($\delta{=}1.001$) & \bf 4.314 & \bf 1760 & \bf
  361 & 3.14 & \bf 512\,/\,512 \\
\midrule
{\bf CelebA-64} & Vanilla VAE                                    & 3.896 & 2198
  & 3283 & 1.00 & 985\,/\,1024 \\
  & $\bm{\lambda}$-\textbf{VAE} ($\delta{=}1.001$) & \bf 3.655 & \bf 3646 & \bf
  6596 & 1.99 & \bf 1024\,/\,1024 \\
\bottomrule
\end{tabular}
\end{table}

\subsection{Qualitative Comparison}
\label{sec:exp_generation}

In this section, we show the impact of increased information flow and minimized
posterior collapse on reconstruction quality.

\begin{figure}[t]
\centering
\begin{subfigure}[t]{1.0\linewidth}
  \centering
  \includegraphics[width=\linewidth]{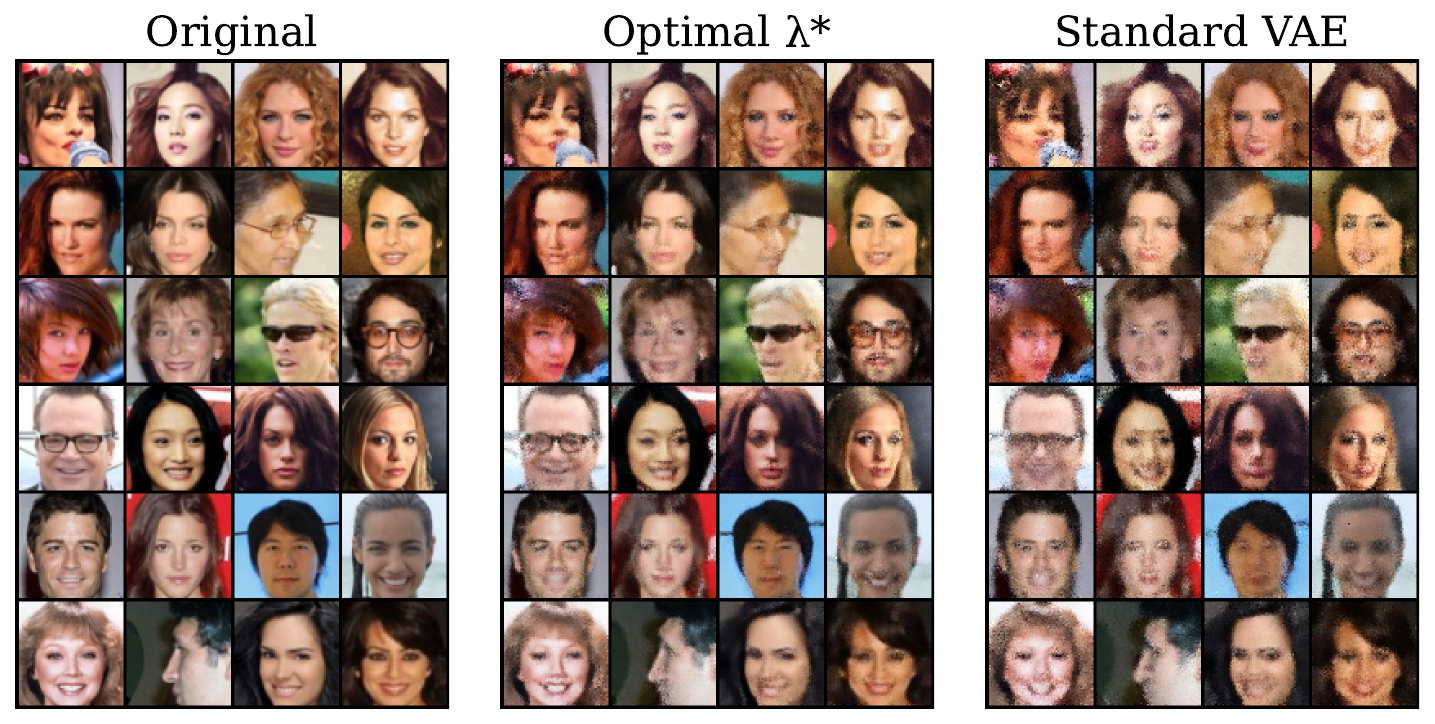}
  \caption{\textbf{CelebA-64}}
  \label{fig:celeb64_reconstruction}
\end{subfigure}
\begin{subfigure}[t]{1.0\linewidth}
  \centering
  \includegraphics[width=\linewidth]{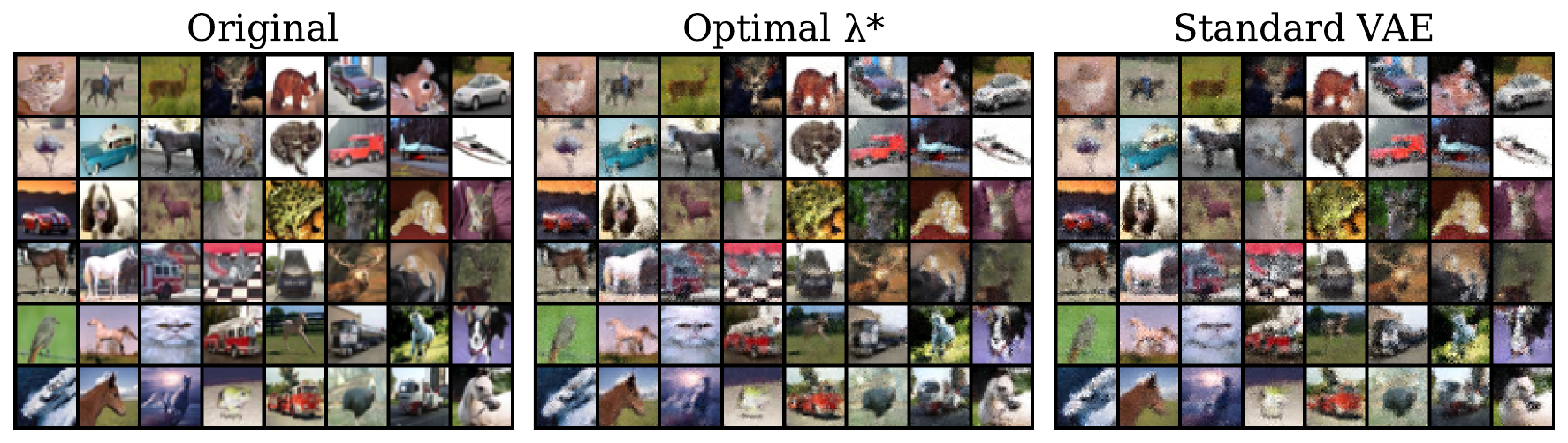}
  \caption{\textbf{CIFAR10}}
  \label{fig:cifar10_reconstruction}
\end{subfigure}
\caption{\textbf{Qualitative Comparison On Reconstructions}.
Optimal $\lambda$ produces sharper, more detailed reconstructions,
consistent with the 2.1× capacity recovery in Table~\ref{tab:image_datasets}.
Differences in quality are best observed zoomed in.}
\label{fig:reconstruction_comparison_inf_flow}
\end{figure}

\begin{figure}[t]
  \centering
  \begin{subfigure}[t]{0.5\linewidth}
  \centering
  \includegraphics[width=\linewidth]{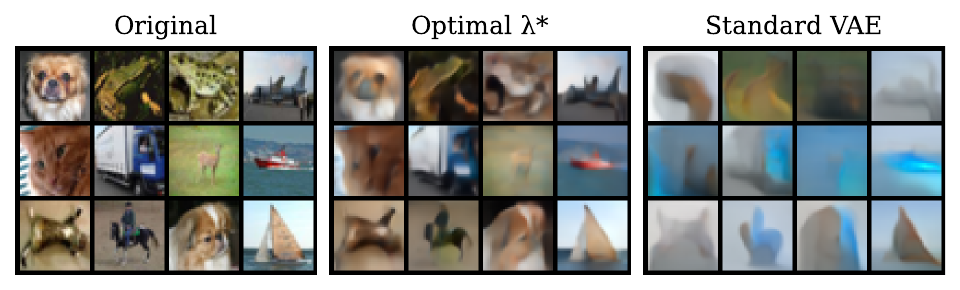}
  \caption{Greedy decoding (argmax at each pixel step).}
  \label{fig:cifar10_greedy}
\end{subfigure}\hfill\hfill
\begin{subfigure}[t]{0.5\linewidth}
  \centering
  \includegraphics[width=\linewidth]{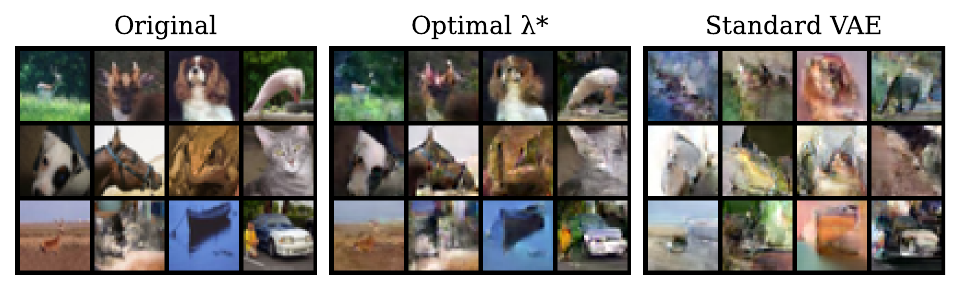}
  \caption{Ancestral sampling from $p(x_i|x_{<i},z)$.}
  \label{fig:cifar10_sampled}
\end{subfigure}
\caption{\textbf{Reconstruction with PixelCNN decoder on CIFAR-10.}
Each panel shows ground-truth (left), Standard VAE (middle), and $\lambda$-VAE
(right) for the same test images.  Despite near-identical ELBO-BPD (3.518 vs.\
3.494, Table~\ref{tab:auto_regression}), the standard VAE has 378/512 collapsed
latent dimensions and decoder capacity of only 12.1 nats: the PixelCNN has
learned to model the image from pixel context alone and ignores $z$.  This
collapse is visible in both sampling modes (greedy and ancestral) where
$\lambda$-VAE reconstructions retain input-specific detail even without teacher
forcing.}
\label{fig:reconstruction_pixelcnn}
\end{figure}

In Figure~\ref{fig:reconstruction_comparison_inf_flow}, we compare
reconstructions between the standard and optimal-$\lambda$ VAE. $\lambda$-VAE
produces sharper images with more coherent pixel values on both CIFAR-10 and
CelebA-64. This qualitative improvement follows directly from the $2.8\times$
gain in SNR information capacity and the $2.1\times$ gain in decoder capacity
shown in Figure~\ref{fig:info_gap_dynamics} and Table~\ref{tab:image_datasets}.
Reconstruction quality degradation due to posterior collapse, however,  
is much more prominently observed in a case where the modelling is primarily
absorbed by the decoder. A sufficiently powerful decoder trained with standard
VAE causes significant gradient imbalance that can render most of the latent
code uninformative and wasted. This phenomenon is more apparent in
autoregressive models like pixelCNN decoders~\cite{gulrajani2016pixelvae} where
the decoder is trained with teacher forcing and can quickly learn to ignore most
of the latent codes, especially pixel level information when it is conditioned
on the ground-truth of previous pixel. Hence, leading to a significant posterior
collapse and poor sample quality during \emph{inference}, despite a comparably
good BPD during training. In contrast, the optimal $\lambda$ keeps $z_i$ close
to  $\mu_i$ with high SNR throughout training with reasonable variance
(depending on $\delta$), ensuring the decoder cannot bypass the latent code
regardless of decoder capacity. We show this contrast by using the same
architecture as before only replacing the decoder with a PixelCNN decoder.  Both
models achieve comparable ELBO-BPD at evaluation (3.518 standard vs.\ 3.494
optimal; Table~\ref{tab:auto_regression}), yet 378 of 512 latent dimensions are
effectively collapsed in the standard VAE versus none in the optimal $\lambda$
model.  Decoder capacity further confirms this; the standard VAE's PixelCNN
allocates only 12.1 nats to the latent code while $\lambda$-VAE allocates 75.0
nats ($6.2{\times}$ more), meaning the autoregressive decoder has learned to
model CIFAR-10 almost entirely from its own context.  The collapse is exposed
when sampling from the model in both \emph{greedy} mode (argmax at each pixel
conditioned on $z$ and all previous pixels) and \emph{ancestral} sampling
($p(x_i|x_{<i},z)$ drawn sequentially per pixel), as shown in
Figure~\ref{fig:reconstruction_pixelcnn}.  This is primarily due to preserved
information flow under optimal $\lambda$-VAE even though teacher forcing was
used in both cases for training.

\section{Conclusion}
\label{sec:conclusion}

Posterior collapse is commonly attributed to the $\mathbb{KL}$ term overwhelming
reconstruction, but this framing misses the mechanism.  The $\mathbb{KL}$ term
does not suppress all dimensions uniformly -- it drives \emph{specific}
dimensions toward $\sigma_i = 1$ through two coupled pathways. At $\sigma_i =
1$, both pathways converge: it is simultaneously a stable fixed point of the
training dynamics where the reconstruction gradient vanishes
(Proposition~\ref{prop:ratio_grad}), and the regime in which injected noise
overwhelms the encoded signal, reducing per-channel SNR and decoupling the
decoder from the latent code (Proposition~\ref{prop:enabling}). $\lambda$-VAE
corrects both by shifting the stable attractor from $\sigma_i = 1$ to $\sigma^*
= \lambda^{-1/(\lambda-1)} < 1$, simultaneously restoring the reconstruction
gradient and calibrating per-channel noise to signal strength. The per-dimension
optimal $\lambda$ estimation (Proposition~\ref{prop:optimal_lambda}) follows
from a net information gain objective that balances information throughput
against marginal-prior divergence, with $\delta$ as the only hyperparameter.
Empirically, $\lambda$-VAE consistently reduces collapsed dimensions, recovers
information throughput, and improves reconstruction quality across all tested
scales -- synthetic, binarized, and large-scale RGB -- with the PixelCNN
experiment providing the sharpest demonstration with near-identical BPD (3.518
vs.\ 3.494) yet $6.2\times$ more decoder capacity allocated to the latent code.

The broader implications for general training principles can be summarized as
follows: a stochastic bottleneck trained end-to-end with a sufficiently powerful
downstream module will be bypassed when the injected noise exceeds the signal
capacity. This is not because the model cannot learn, but because bypassing is
the lower-cost solution under a uniform noise budget. $\lambda$-VAE instantiates
a targeted correction for Gaussian posteriors, but the same principle is
expected to apply wherever the noise level at a stochastic layer is set globally
rather than calibrated to per-channel signal strength. Three directions are left
for future work. \emph{Hierarchical VAEs:} collapse at early latent levels
propagates through the hierarchy; per-level $\lambda^*$ schedules may provide
targeted repair without the careful $\mathbb{KL}$ balancing that hierarchical
training currently requires. \emph{Adaptive $\delta$:} the current approach sets
$\delta$ once per experiment; learning $\delta$ or scheduling it based on the
evolving $\sigma$ distribution could recover the reconstruction, generation
tradeoff automatically, and reduce per-dataset tuning. \emph{Beyond VAEs:} the
information gap--marginal mismatch duality (Eq.~\ref{eq:duality}) provides a
cheap, dataset-free proxy for aggregate posterior quality that avoids the
integration cost of InfoVAE or Wasserstein objectives; whether this proxy can
guide training in discrete bottleneck models or diffusion-based posteriors is an
open question.

\begin{table}[t]
\centering
\caption{\textbf{PixelCNN decoder comparison on CIFAR-10.}
BPD evaluated via the test-set ELBO. Decoder capacity sums per-dimension
contributions from the training history (a proxy for how much the decoder relies
on $z$).}
\label{tab:auto_regression}
\small
\begin{tabular}{lrrrr}
\toprule
\textbf{Model}
  & \textbf{BPD}$\downarrow$
  & \textbf{AU}$\uparrow$
  & \textbf{Dec.\ cap.\ (nats)}$\uparrow$
  & $\bar{\lambda}$ \\
\midrule
Standard VAE              & 3.518          & 134\,/\,512          & 12.1         & 1.00 \\
$\lambda$-VAE ($\delta{=}1.001$) & \textbf{3.494} & \textbf{512\,/\,512} & \textbf{75.0} & 6.37 \\
\bottomrule
\end{tabular}
\end{table}




\bibliography{main}
\bibliographystyle{tmlr}

\appendix
\section{Appendix}
\addcontentsline{toc}{section}{Appendix~A: Gradient Imbalance and Information gap.}
 
\subsection*{A.1\quad Gradient Imbalance (Proposition~\ref{prop:ratio_grad})}
\label{ap:gradient_imbalance}

The reconstruction gradient w.r.t.\ $\sigma_i$ is defined as
\begin{equation}
  g_i^\sigma \;:=\; \frac{\partial R}{\partial \sigma_i}
    \;=\; \mathbb{E}_{p(\mathbf{x}),\,\epsilon_i}\!\left[
            \nabla_{z_i}\log p_\theta(\mathbf{x}|\mathbf{z})\cdot\epsilon_i
          \right],
  \label{eq:recon_grad}
\end{equation}
while the $\mathbb{KL}$ gradient is $\partial \mathbb{KL}_i/\partial \sigma_i = \sigma_i - 1/\sigma_i$.
Combining, the ELBO gradient is 
\begin{equation}
  \frac{\partial\mathcal{L}}{\partial\sigma_i}
    \;=\; g_i^\sigma + \frac{1}{\sigma_i} - \sigma_i. \tag{A.1}\label{eq:ap_elbo_grad}
\end{equation}
From the above, we see the $\mathbb{KL}$ restoring force $1/\sigma_i - \sigma_i$
is positive for $\sigma_i < 1$, and negative for $\sigma_i > 1$. It drives
$\sigma_i \to 1$ whenever it dominates the reconstruction gradient $g_i^\sigma$.

\paragraph{Gradient ratio and fixed points:} Setting the ELBO gradient Eq.~\ref{eq:ap_elbo_grad} to zero
gives the following equilibrium condition:
\begin{equation}
  \sigma_i^* \;=\; \frac{-g_i^\sigma + \sqrt{(g_i^\sigma)^2 + 4}}{2}
    \;\in\; (0,1), \tag{A.2}\label{eq:active_eq}
\end{equation}

When $g_i^\sigma = 0$, this gives the unique
positive root $\sigma_i^* = 1$. Linearising around $\sigma_i = 1 + \delta$ and
simplifying $\partial\mathcal{L}/\partial\sigma_i$ yields $1/\sigma_i - \sigma_i
\approx -2\delta + O(\delta^2)$, so $\partial\mathcal{L}/\partial\sigma_i
\approx -2\delta$. Thus, the collapse fixed point $\sigma_i^* = 1$ is a
\emph{locally asymptotically stable point with linearisation rate~$-2$}.
For $g_i^\sigma > 0$, the gradient ratio is the necessary and sufficient collapse indicator stated in Proposition~\ref{prop:ratio_grad}.
\begin{equation}
  \rho_i \;=\; \frac{|g_i^\sigma|}{|1/\sigma_i - \sigma_i|}. \tag{A.3}\label{eq:rho_def}
\end{equation}
 
\paragraph{Proof of sufficiency:}
 Assume $\rho_i(t) < 1$ for all $t \geq T_0$. Then $|g_i^\sigma| < |1/\sigma_i -
\sigma_i|$, so $\partial\mathcal{L}/\partial\sigma_i > 0$ whenever $\sigma_i <
1$. The sequence $\{\sigma_i(t)\}_{t \geq T_0}$ is monotonically non-decreasing
and bounded above by~$1$. By the \emph{Monotone Convergence Theorem} it
converges to a limit $\sigma_i^\infty \leq 1$. If $\sigma_i^\infty < 1$, then
$1/\sigma_i^\infty - \sigma_i^\infty > 0$, and Eq.~\ref{eq:ap_elbo_grad}
requires $g_i^\sigma(\sigma_i^\infty) = 0$ at a fixed point, which forces
$\sigma_i^\infty = 1$, which is a contradiction. Hence $\sigma_i^\infty =
1$. \hfill$\blacksquare$
 
\paragraph{Proof of necessity:} Under the decay condition provided in
the proposition, near $\sigma_i = 1$ the $\mathbb{KL}$ part is written as
\begin{equation}
  \left|\frac{1}{\sigma_i} - \sigma_i\right|
    \;=\; \frac{(1-\sigma_i)(1+\sigma_i)}{\sigma_i}
    \;\sim\; 2(1-\sigma_i)
    \quad\text{as } \sigma_i \to 1^-, \tag{A.4}
    \label{eq:kl_asym}
\end{equation}
%
while the decay bound $|g_i^\sigma| = O\!\left((1-\sigma_i)^\alpha\right)$ gives
the ratio bounded by
\begin{equation}
  \rho_i \;=\; \frac{O\!\left((1-\sigma_i)^\alpha\right)}{2(1-\sigma_i)}
    \;=\; O\!\left((1-\sigma_i)^{\alpha-1}\right). \tag{A.5}\label{eq:rho_asym}
\end{equation}
\begin{itemize}
  \item \textbf{$\alpha > 1$:}\;
        $(1-\sigma_i)^{\alpha-1} \to 0$ as $\sigma_i \to 1^-$, so $\rho_i \to 0$.
        Collapse implies $\rho_i \to 0$. Hence, \emph{The ratio is a \textbf{necessary and sufficient}
        collapse signal for dimension $i$}. \hfill$\blacksquare$
 
  \item \textbf{$\alpha = 1$:}\;
        $\rho_i \to g_0/2 > 0$ as $\sigma_i \to 1$: the ratio stays bounded away
        from zero regardless of whether collapse occurs, giving a potential \emph{false negative}.
        The condition $\alpha > 1$ in the proposition is therefore not a restriction but a
        \emph{regime identifier}; it characterises the decoders for which $\rho_i$ is
        an informative diagnostic.
 
  \item \textbf{$\alpha < 1$:}\;
        $\rho_i \to \infty$; the decoder resists collapse. No collapse occurs.
\end{itemize}

\subsection*{A.2\quad Information Gap (Proposition~\ref{prop:enabling})}
\label{app:information_gap}
 
We derive the gap $\Delta I = I(X;\phi) - I(X;Z) \geq 0$ under Gaussian
posteriors and show it attenuates change in the encoder.
By the
\emph{Data Processing Inequality}~\cite{cover1999elements}, we have
\begin{equation}
  I(X;Z) \;\leq\; I(X;\phi)
    \qquad\Longrightarrow\qquad
    \Delta I \;=\; I(X;\phi) - I(X;Z) \;\geq\; 0.
  \label{eq:dpi_bound}
\end{equation}
%
Since $z_i = \mu_i + \sigma_i\epsilon_i$ with $\sigma_i > 0$ is not invertible
(given $z_i = z$, we cannot recover $\epsilon_i$ without $\mu_i, \sigma_i$),
the map is strictly lossy: $\Delta I > 0$ for all $\sigma_i > 0$.
Equivalently, by the chain rule for mutual information, the gap satisfies
\begin{equation}
  \Delta I \;=\; I\!\left(X;\phi\mid Z\right) \;\geq\; 0, \tag{A.6}\label{eq:gap_cmi}
\end{equation}
the conditional mutual information between~$X$ and the encoder parameters given~$Z$.
 
\paragraph{Information gap under Gaussian posteriors:} Consider an approximate 
posterior estimate $q(z_i|x) = \mathcal{N}(\mu_i(x),\sigma_i^2(x))$, and its
conditional entropy
\begin{equation}
  H[Z_i|X]
    \;=\; \tfrac{1}{2}\log(2\pi e)
        + \tfrac{1}{2}\mathbb{E}_{p(x)}\!\left[\log\sigma_i^2(x)\right].
  \tag{A.7}\label{eq:cond_ent}
\end{equation}
Since $I(X;\phi) = H[\phi]$ is fixed by the encoder weights (encoder is deterministic),
changes in the gap $\Delta I$ are driven by changes in $I(X;Z)$.
Under approximate marginal matching $q(z) \approx \mathcal{N}(\mathbf{0},\mathbf{I})$
(valid under optimized $\mathbb{KL}$ regularization),
the accessible mutual information reduces to
\begin{equation}
  I(X;Z)
    \;\approx\; \frac{1}{2}\sum_{i=1}^{d}
      \mathbb{E}_{p(x)}\!\left[-\log\sigma_i^2(x)\right]
    \;\geq\; 0.
  \tag{A.8}\label{eq:gap_approx}
\end{equation}
The gap grows monotonically: $\partial(\Delta I)/\partial\sigma_i = 1/\sigma_i > 0$.
At collapse ($\sigma_i \to 1$), dimension~$i$'s gap reaches its maximum $I(X;\phi_i)$
while $I(X;Z_i) \approx -\tfrac{1}{2}\log\sigma_i^2 \to 0$,
reflecting complete information loss.
 
\paragraph{Derivation of the sensitivity equation:}
 
From the definition, we have the identity
\begin{equation}
  I(X;Z) \;=\; I(X;\phi) - \Delta I. \tag{A.9}\label{eq:ixz_identity}
\end{equation}
Treating $\Delta I$ as a function of $I(X;\phi)$  and differentiating
Eq.~\ref{eq:ixz_identity} along the training trajectory
\begin{equation}
  \frac{\partial\,I(X;Z)}{\partial\,I(X;\phi)}
    \;=\; 1 - \frac{\partial\,\Delta I}{\partial\,I(X;\phi)}.
  \tag{A.10}\label{eq:ixz_deriv}
\end{equation}
Since $\mathcal{R} = f(I(X;Z))$ for monotonically increasing~$f$, the chain rule gives
\begin{equation}
  \frac{\partial\mathcal{R}}{\partial\,I(X;\phi)}
    \;=\; \frac{\partial\mathcal{R}}{\partial\,I(X;Z)}
          \cdot\left(1 - \frac{\partial\,\Delta I}{\partial\,I(X;\phi)}\right).
  \tag{A.11}\label{eq:sensitivity}
\end{equation}
This is equation Eq.~\ref{eq:enabling} in the main text.

\paragraph{Duality with marginal mismatch:}

From the standard ELBO decomposition \citep{hoffman2016elbo},
\begin{align}
   \mathbb{KL}(q(z|x)\|p(z))
    \;=\; I(X;Z) + \mathbb{KL}(q(z)\|p(z)).
  \tag{A.12}
  \label{eq:elbo_decomp}
\end{align}
Substituting Eq.~\ref{eq:ixz_identity}:
\begin{align}
  \mathbb{KL}(q(z)\|p(z)) \;=\; \mathbb{KL}(q(z|x)\|p(z)) - I(X;\phi) + \Delta I. \tag{A.13}\label{eq:m_gap}
\end{align}


\section*{Appendix~B}
\addcontentsline{toc}{section}{Appendix~B: Variance equalization and optimal $\lambda$}
\label{app:optimal_lambda_derivation}

\subsection*{B.1\quad Equilibrium Variance Target
            $\sigma^* = \lambda^{-1/(\lambda-1)}$}
\label{app:sigma_star}

Under $\lambda$-scaling, the reparameterization is $z_i = \mu_i +
\sigma_i^\lambda\epsilon_i$. Differentiating the ELBO objective with respect to
$\log\sigma_i$, the modification introduced by $\lambda$-scaling relative to
standard sampling produces the following scaling term 
\begin{align}
  \Delta\nabla_i := \frac{\partial \mathcal{L}_\lambda}{\partial \log\sigma_i}
  -
  \frac{\partial \mathcal{L}_1}{\partial \log\sigma_i}
  \;=\;
  \bigl(\lambda\,\sigma_i^\lambda - \sigma_i\bigr)\cdot g_i^\sigma,
  \label{eq:delta_grad}
\end{align}
where $g_i^\sigma := \mathbb{E}_{\epsilon_i}[\nabla_{z_i}\log p_\theta(x|z)\cdot\epsilon_i]$
is the reconstruction gradient.
The derivation follows from the chain rule through $\sigma_i^\lambda$:
$\partial(\sigma_i^\lambda\epsilon_i)/\partial\log\sigma_i = \lambda\sigma_i^\lambda$,
compared to $\partial(\sigma_i\epsilon_i)/\partial\log\sigma_i = \sigma_i$ in
the standard case.

The scaling term in Eq.~\ref{eq:delta_grad} is zero iff $\lambda\sigma_i^\lambda = \sigma_i$, which gives
the equilibrium condition for $\sigma_i$:

\begin{align}
  \lambda\,\sigma_i^{\lambda-1} &\;=\; 1 \nonumber \\
  \sigma_i^* &\;=\; \lambda^{-1/(\lambda-1)}.
  \label{eq:sigma_star_derived}
\end{align}
The above is valid for $\lambda > 1$.

\paragraph{$\sigma^*$ is a stable attractor:}
The scaling term Eq.~\ref{eq:delta_grad} acts as a restoring force.
For $g_i^\sigma > 0$, the following conditions hold
\begin{itemize}
  \item $\sigma_i < \sigma^*$: then $\sigma_i^{\lambda-1} < 1/\lambda$,
        so $\lambda\sigma_i^\lambda < \sigma_i$, giving
        $\Delta\nabla_i < 0$.  The gradient on $\log\sigma_i$ is negative:
        $\sigma_i$ is pushed \emph{upward} toward $\sigma^*$.

  \item $\sigma_i > \sigma^*$: then $\sigma_i^{\lambda-1} > 1/\lambda$,
        so $\lambda\sigma_i^\lambda > \sigma_i$, giving
        $\Delta\nabla_i > 0$.  The gradient on $\log\sigma_i$ is positive:
        $\sigma_i$ is pushed \emph{downward} toward $\sigma^*$.
\end{itemize}
The force is bidirectional and vanishes only at $\sigma^*$: $\sigma^*$ is a
stable fixed point. Every dimension, regardless of its initial value, is driven
toward the \emph{same} target. See Figure~\ref{fig:equalization_convergence} and
Figure~\ref{fig:equalization_exp} for illustrative and empirical data of this
behaviour.

\subsection*{B.2\quad Optimal Per-Dimension $\lambda^*_i$}
\label{app:lambda_star}

\paragraph{Objective:}
For a single latent dimension~$i$ with encoder variance $\sigma_i \in (0,1)$,
the per-dimension net information gain objective
(Eq.~\ref{eq:objective} in the main text) is:
\begin{equation}
  \mathcal{J}(\lambda_i, \sigma_i)
  \;=\;
  (\lambda_i - 1)\,|\!\log\sigma_i|
  \;-\;
  \delta\,\mathbb{KL}(\lambda_i),
  \qquad \delta > 1,
  \label{eq:objective_app}
\end{equation}
where $|\log\sigma_i| = -\log\sigma_i > 0$ for $\sigma_i \in (0,1)$, and
the marginal $\mathbb{KL}$ cost (Eq.~\ref{eq:mismatch} in the main text) is:
\begin{equation}
  \mathbb{KL}(\lambda_i)
  \;=\; \frac{1}{2}\!\left(
    \sigma_i^{2\lambda_i} - \sigma_i^2
    - \log\sigma_i^{2\lambda_i} + \log\sigma_i^2
  \right).
  \label{eq:M_app}
\end{equation}

\paragraph{Derivative of $\mathcal{J}$ with respect to $\lambda_i$:}

The information gain term contributes $|\log\sigma_i|$, which is constant in
$\lambda_i$. For the marginal cost, the only $\lambda_i$-dependent terms
in Eq.~\ref{eq:M_app} are $\sigma_i^{2\lambda_i}$ and $\log\sigma_i^{2\lambda_i} =
2\lambda_i\log\sigma_i$. Hence,
\begin{align}
  \frac{\partial\mathbb{KL}}{\partial\lambda_i}
  &\;=\; \frac{1}{2}\!\left(
    2\sigma_i^{2\lambda_i}\log\sigma_i
    - 2\log\sigma_i
  \right)
  \;=\; \log\sigma_i\!\left(\sigma_i^{2\lambda_i} - 1\right).
  \label{eq:dM_dlam}
\end{align}
For $\sigma_i \in (0,1)$: $\log\sigma_i < 0$ and $\sigma_i^{2\lambda_i} < 1$, so
$\sigma_i^{2\lambda_i} - 1 < 0$. Therefore
$\partial\mathbb{KL}/\partial\lambda_i = \log\sigma_i(\sigma_i^{2\lambda_i}-1) >
0$: the marginal cost is strictly increasing in $\lambda_i$. The full derivative
of the objective is
\begin{equation}
  \frac{\partial\mathcal{J}}{\partial\lambda_i}
  \;=\; |\log\sigma_i|
        - \delta\log\sigma_i\!\left(\sigma_i^{2\lambda_i} - 1\right).
  \label{eq:dJ_dlam}
\end{equation}

\paragraph{Stationary-point condition:}

Setting Eq.~\ref{eq:dJ_dlam} to zero,
\begin{equation}
  |\log\sigma_i|
  \;=\; \delta\log\sigma_i\!\left(\sigma_i^{2\lambda_i} - 1\right).
  \label{eq:foc}
\end{equation}
Substituting $|\log\sigma_i| = -\log\sigma_i$ (since $\sigma_i < 1$):
\begin{equation}
  -\log\sigma_i
  \;=\; \delta\log\sigma_i\!\left(\sigma_i^{2\lambda_i} - 1\right).
  \label{eq:foc_sub}
\end{equation}
Dividing both sides by $\log\sigma_i \neq 0$:
\begin{equation}
  -1 \;=\; \delta\!\left(\sigma_i^{2\lambda_i} - 1\right).
  \label{eq:foc_div}
\end{equation}
Rearranging,
\begin{equation}
  \sigma_i^{2\lambda_i} \;=\; 1 - \frac{1}{\delta}.
  \label{eq:sigma_power}
\end{equation}

\paragraph{Solving for $\lambda_i^*$:}

Taking the natural logarithm of both sides of Eq.~\ref{eq:sigma_power}:
\begin{equation}
  2\lambda_i^*\,\log\sigma_i \;=\; \log\!\left(1 - \frac{1}{\delta}\right).
  \label{eq:log_eq}
\end{equation}
Solving for $\lambda_i^*$:
\begin{equation}
  \lambda_i^*
  \;=\; \frac{\log(1 - 1/\delta)}{2\log\sigma_i}.
  \label{eq:lambda_star_unconstrained}
\end{equation}
For $\sigma_i \in (0,1)$: $\log\sigma_i < 0$.
For $\delta > 1$: $1 - 1/\delta \in (0,1)$, so $\log(1-1/\delta) < 0$.
Therefore $\lambda_i^* > 0$. Enforcing the constraint $\lambda_i \geq 1$:
\begin{equation}
  \boxed{\lambda_i^*
  \;=\; \max\!\left(1,\;\frac{\log(1 - 1/\delta)}{2\log\sigma_i}\right).}
  \label{eq:lambda_star_derived}
\end{equation}

\paragraph{The stationary point is a global maximum:}

The second derivative of Eq.~\ref{eq:objective_app}:
\begin{equation}
  \frac{\partial^2\mathcal{J}}{\partial\lambda_i^2}
  \;=\; -\delta\,\frac{\partial^2\mathbb{KL}}{\partial\lambda_i^2}.
  \label{eq:second_deriv}
\end{equation}
From Eq.~\ref{eq:dM_dlam}:
\begin{equation}
  \frac{\partial^2\mathbb{KL}}{\partial\lambda_i^2}
  \;=\; \frac{\partial}{\partial\lambda_i}
        \!\left[\log\sigma_i\!\left(\sigma_i^{2\lambda_i} - 1\right)\right]
  \;=\; 2\,(\log\sigma_i)^2\,\sigma_i^{2\lambda_i}
  \;>\; 0.
  \label{eq:M_second}
\end{equation}
Therefore:
\begin{equation}
  \frac{\partial^2\mathcal{J}}{\partial\lambda_i^2}
  \;=\; -2\delta\,(\log\sigma_i)^2\,\sigma_i^{2\lambda_i}
  \;<\; 0.
  \label{eq:concave}
\end{equation}
The objective is strictly concave in $\lambda_i$, confirming that the
stationary point Eq.~\ref{eq:lambda_star_derived} is a global maximum. Hence, 
maximizing the net information gain for a given $\delta$ hyperparameter. 

\end{document}